\documentclass[review]{elsarticle}

\makeatletter
\def\ps@pprintTitle{%
 \let\@oddhead\@empty
 \let\@evenhead\@empty
    \def\@oddfoot{\footnotesize\itshape
         {Published as a conference paper at AISTATS 2023} \hfill\today}%
 \let\@evenfoot\@oddfoot}
\makeatother

\usepackage{amsmath}
\usepackage{amsfonts}
\usepackage{amssymb}
\usepackage{amsthm}
\usepackage{bm}
\usepackage{indentfirst}
\usepackage{graphicx}
\usepackage{subfigure}
\usepackage{array}
\usepackage{fullpage}

\newcommand{\mb}{\mathbf}

\DeclareMathAlphabet{\mathsfsl}{OT1}{cmss}{m}{sl}

\newcommand{\PreserveBackslash}[1]{\let\temp=\\#1\let\\=\temp}
\newcolumntype{C}[1]{>{\PreserveBackslash\centering}p{#1}}
\newcolumntype{R}[1]{>{\PreserveBackslash\raggedleft}p{#1}}
\newcolumntype{L}[1]{>{\PreserveBackslash\raggedright}p{#1}}

\numberwithin{equation}{section}

\theoremstyle{definition}

\usepackage{anyfontsize}
\usepackage{enumerate}
\usepackage{esint}
\usepackage{etoolbox}
\usepackage{isomath}
\usepackage{mathrsfs}
\usepackage{mathtools}
\usepackage{multicol}
\usepackage{pgfplots}
\usepackage{scalerel}
\usepackage{stackengine}
\usepackage{tabulary}
\usepackage{tikz}

\makeatletter
\newcommand*\bdot{\mathpalette\bdot@{.65}}
\newcommand*\bdot@[2]{\mathbin{\vcenter{\hbox{\scalebox{#2}{$\m@th#1\bullet$}}}}}
\makeatother

\makeatletter
\newcommand*\bddot{\mathpalette\bddot@{.65}}
\newcommand*\bddot@[2]{\mathbin{\vcenter{\hbox{\scalebox{#2}
    {$\m@th#1\smash{{}_{\bullet}^{\bullet}}$}}}}}
\makeatother

\newcommand{\circled}[2][]{%
  \tikz[baseline=(char.base)]{%
    \node[shape = circle, draw, inner sep = .5pt]
    (char) {\phantom{\ifblank{#1}{#2}{#1}}};%
    \node at (char.center) {\makebox[0pt][c]{#2}};}}
\robustify{\circled}

\stackMath
\newcommand\reallywidecheck[1]{%
\savestack{\tmpbox}{\stretchto{%
  \scaleto{%
    \scalerel*[\widthof{\ensuremath{#1}}]{\kern-.6pt\bigwedge\kern-.6pt}%
    {\rule[-\textheight/2]{1ex}{\textheight}}%WIDTH-LIMITED BIG WEDGE
  }{\textheight}%
}{0.5ex}}%
\stackon[1pt]{#1}{\scalebox{-1}{\tmpbox}}%
}
\usepackage{amsmath,amssymb,amsfonts,mathrsfs}
\usepackage{hyperref}
\usepackage{url}
\usepackage{bm}
\usepackage{amsthm}
\biboptions{sort&compress}

\usepackage{graphicx}
\graphicspath{{figures/}}

\newcommand{\real}{\mathbb{R}}

\newcommand{\mcS}{\mathcal{S}}

\newcommand{\mcD}{\mathcal{D}}

\newcommand{\mcG}{\mathcal{G}}

\newcommand{\mcU}{\mathcal{U}}
\newcommand{\mbF}{\mathbb{F}}
\newcommand{\mbU}{\mathbb{U}}

\newcommand{\mcL}{\mathcal{L}}

\newcommand{\mcT}{\mathcal{T}}

\newcommand{\mcP}{\mathcal{P}}
\newcommand{\mcQ}{\mathcal{Q}}

\usepackage{xcolor}

\def\omg{{\Omega}}

\def\omgb{\mathcal{B}\Omega}
\def\omgbb{\mathcal{B}\mathcal{B}\Omega}

\def \fb{\bm{f}}

\def \ab{\bm{a}}

\def \ub{\bm{u}}

\def \xb{\bm{x}}
\def \mb{\bm{m}}

\def \hb{\bm{h}}

\newcommand{\vertii}[1]{{\left\vert\left\vert #1
    \right\vert\right\vert}}    
    
\newcommand{\verti}[1]{{\left\vert #1
    \right\vert}}  
    
\newcommand{\YY}[2][black]{{\textcolor{#1}{#2}}}

\newtheorem{theorem}{Theorem}

\begin{document}

\begin{frontmatter}

\title{INO: Invariant Neural Operators for Learning Complex\\ Physical Systems with Momentum Conservation}

\address[nl]{Global Engineering and Materials, Inc., 1 Airport Place, Princeton, NJ 08540, USA}
\address[yy]{Department of Mathematics, Lehigh University, Bethlehem, PA 18015, USA}

\author[nl]{Ning Liu}\ead{ningliu@umich.edu}
%\author[cl]{Colton J. Ross}\ead{cjross@ou.edu}
%\author[cl]{Chung-Hao Lee}\ead{ch.lee@ou.edu}
\author[yy]{Yue Yu\corref{cor1}}\ead{yuy214@lehigh.edu}
\cortext[cor1]{Corresponding author}
\author[yy]{Huaiqian You}\ead{huy316@lehigh.edu}
\author[yy]{Neeraj Tatikola}\ead{nkt222@lehigh.edu}

\begin{abstract}
Neural operators, which emerge as implicit solution operators of hidden governing equations, have recently become popular tools for learning responses of complex real-world physical systems. Nevertheless, the majority of neural operator applications has thus far been data-driven, which neglects the intrinsic preservation of fundamental physical laws in data. In this paper, we introduce a novel integral neural operator architecture, to learn physical models with fundamental conservation laws automatically guaranteed. In particular, by replacing the frame-dependent position information with its invariant counterpart in the kernel space, the proposed neural operator is by design translation- and rotation-invariant, and consequently abides by the conservation laws of linear and angular momentums. As applications, we demonstrate the expressivity and efficacy of our model in learning complex material behaviors from both synthetic and experimental datasets, and show that, by automatically satisfying these essential physical laws, our learned neural operator is not only generalizable in handling translated and rotated datasets, but also achieves state-of-the-art accuracy and efficiency as compared to baseline neural operator models. %Therefore, our INO can serve as a more efficient and robust model for learning real-world mechanical systems.
\end{abstract}


\end{frontmatter}

\section{Introduction}

\begin{figure}[!t]\centering
    \includegraphics[width=.55\linewidth]{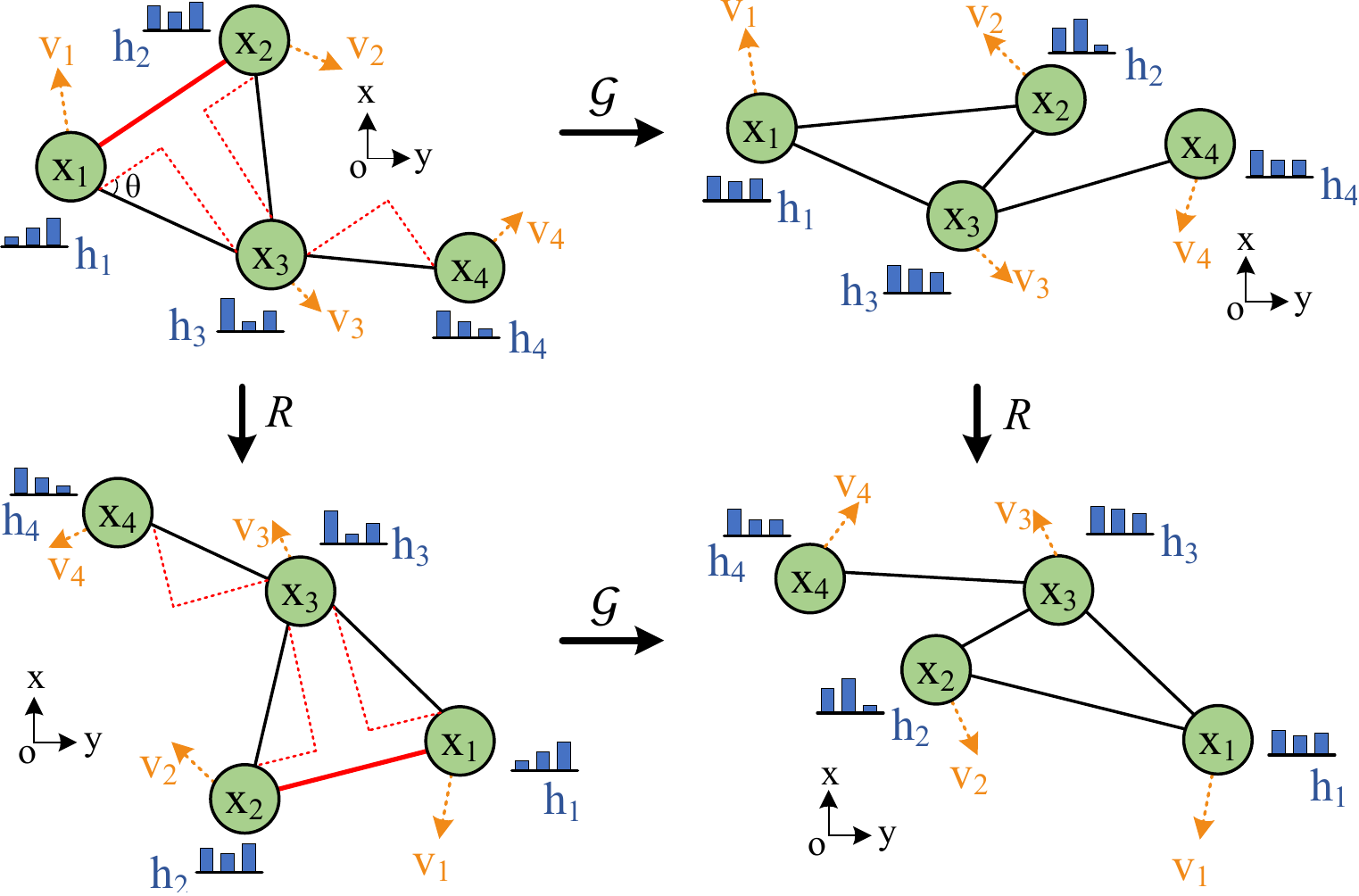}
    \caption{An illustration of the translational and rotational invariance on a mesh grid with INO. Here, $\mcG$ represents the learnt mapping, $R$ represents the rotation and translation of coordinate frames.  To design an invariant neural operator, our key idea is to characterize the interaction between nodes via an invariant kernel. The solid red lines represent the selected local reference edges and the dashed red lines indicate the two components of relative Euclidean distance, $\overline{x_j-x_i}:=[\verti{x_j-x_i}\cos\theta,\verti{x_j-x_i}\sin\theta]$, with which we parameterize the proposed kernel form.}
    \label{fig:equivariance}
\end{figure}

%\vspace{-10pt}

Neural operators \citep{anandkumar2020neural, li2020fourier,lu2019deeponet} have gained popularity in recent years as a form of implicit solution operators to unveil hidden physics of complex real-world physical systems from data. Benefiting from their integral form of the architecture, neural operators {learn a surrogate mapping between function spaces, which} are resolution independent and can be generalized to different input instances \citep{kovachki2021neural}. Resolution independence empowers the learned operator to retain consistent accuracy in prediction regardless of the variation of input resolutions, while being generalizable to different input instances offers the possibility to solve unseen input instances with only a forward pass of the trained network without the hassle of repeating the time-consuming training process. These facts make neural operators excellent candidates for providing surrogate models for complex physical systems \citep{li2021physics,goswami2022pino}. 

Despite of the notable advances in the development of neural operators over traditional neural networks (NNs), their performance \YY{highly rely on the amount of available data}, especially when the governing PDE is unknown \cite{goswami2022pino}. An effective way to alleviate this pathology is to incorporate into the designed architecture the intrinsic preservation of fundamental physical laws in data, as is the case of the conservation of linear and angular momentums in most physical systems. In \cite{you2022physics}, it was found that incorporating partial physics, such as the no-permanent-set assumption, can enhance the prediction of neural operators in out-of-distribution prediction tasks. However, to the authors' best knowledge, neural operators that preserve the conservation laws of linear and angular momentums have not yet been explored.

To enhance the accuracy of neural operators in physical system predictions, in this work we propose the Invariant Neural Operator (INO), a novel integral neural operator architecture that remains invariant under frame translations and rotations. %As such, INO guarantees the conservation of linear and angular momentums \citep{noether1971invariant,sarlet1981generalizations}. 
Specifically, we substitute the frame-dependent coordinate information with its invariant counterpart 
%(i.e., the decomposed Euclidean distance between nodes computable by the edge length and local orientation, 
(cf. Figure~\ref{fig:equivariance}) in the kernel space, so that the learned kernel is independent of geometric transformations. 
%, and consequently abides by the conservation laws of linear and angular momentums. 
Compared to existing neural operator methods, the proposed architecture mainly carries four significant advantages. First, different from existing physics-informed neural operators, our approach only requires observed data pairs and does not rely on known governing equations \citep{goswami2022physics,goswami2022pino,li2021physics,wang2021learning}. Therefore, it is readily applicable to learn physical systems directly from experimental measurements \citep{ranade2021generalized} or simulation data \citep{kim2019peri}, for which the underlying governing equations may not be available. 
Second, the invariant properties in our approach are realized through built-in kernel functions, which is anticipated to be more robust and efficient than data augmentation techniques \citep{quiroga2018revisiting}. 
%it is more robust and efficient to learn an invariant kernel function. 
%Second, since the learned kernel is not affected by frame locations and orientations, it provides more physically consistent model interpretation from data. 
%significantly simplifies data interpretation. 
%More importantly, the decomposed Euclidean distance naturally encodes the geometric information that is predictable under frame transformations, which further simplifies the learning process. 
More importantly, through embedding the invariant kernel and updating frame-dependent coordinate information with a separate network, our architecture naturally stems from the interpretation of its layer update as a particle-based method \citep{karplus2002molecular,liu2010smoothed}, which significantly simplifies model interpretation. Last but not least, analogous to the E(n) equivariance concept in \cite{satorras2021n}, our architecture is not limited to two- or three-dimensional invariance. As a matter of fact, it can be easily scaled up to higher dimensions.
%In addition to the translation- and rotation-invariant design of the architecture, the proposed integral operator is also cast in the framework of shallow-to-deep learning \citep{haber2018learning}, where the parameters of the deeper networks are initialized with the optimal parameters learned on shallower networks. It has been reported in \cite{you2022nonlocal, you2022learning} that the shallow-to-deep initialization technique can accelerate the learning process and lead to stable predictions as the network layers get deep. 
In summary, the contributions of our work are:
\vspace{-0.1in}
\begin{itemize}
\item{We propose INO, a novel integral neural operator architecture that is translation- and rotation-invariant, to learn complex physical systems with guaranteed conservation of linear and angular momentums.}
%\item{We equip the proposed INO with the shallow-to-deep initialization technique to accelerate learning and obtain stable predictions as the depth of the network proliferates.}
\vspace{-0.05in}
\item{Equipped with the shallow-to-deep initialization technique and a coordinate embedding network, our INO finds a physical interpretation from a particle-based method, and obtains stable predictions as the network proliferates in depth.}
\vspace{-0.05in}
\item{Our approach only requires data pairs and does not rely on \textit{a priori} domain knowledge, while the guaranteed momentum conservation laws improve the learning efficacy, especially in small data regime.}
\vspace{-0.25in}
\item{We demonstrate the expressivity and generalizability of INO across a variety of synthetic and real-world experimental datasets, and show that our learned neural operator is not only generalizable in handling translated and rotated datasets, but also provides improved prediction from the baseline neural operators.}
\end{itemize}

\section{Background and Related Work}\label{sec:back}

In this section, we briefly introduce the concept of translational and rotational invariance in classical mechanics, and present its connection to the laws of momentum conservation. Moreover, we review relevant concepts of invariance/equivariance and hidden physical system learning with NNs, which will later become complementary to the proposed INO definition.

Throughout this paper, we use lower case letters to denote vectors, upper case letters for matrices, bold letters for functions, calligraphic letters for operators, and blackboard-bold letters for Banach spaces. For any vector $v$, we use $\verti{v}$ to denote its $l^2$ norm. For any function $\fb$ taking values at nodes $\chi :=\{x_1,x_2,\dots,x_M\}$, $\vertii{\fb}$ denotes its $l^2$ norm, i.e., $\vertii{\fb}:=\sqrt{\sum_{i=1}^M(\fb(x_i))/M}$. $\real^d$ represents the dimension-$d$ Euclidean space.

\subsection{Invariance, Equivariance, and Momentum Conservation Laws}

\vspace{-0.05in}

We consider the learning of complex physical responses of a mechanical system, based on a number of observations of the loading field $\fb_i(x)\in\mbF(\omg;\real^{d_f})$ and the corresponding physical system response $\ub_i(x)\in\mbU(\omg;\real^{d_u})$. Here, $i$ denotes the sample index, $\omg\in\real^d$ is the bounded domain of interests, and $\mbF$ and $\mbU$ describe the Banach spaces of functions taking values in $\real^{d_f}$ and $\real^{d_u}$, respectively. To model the physical responses of such a system, we aim to learn a surrogate operator $\mcG:\mbF\rightarrow \mbU$, that maps the input function $\fb(x)$ to the output function $\ub(x)$. 

Let $\mcT_g:\mbF\rightarrow\mbF$ be a set of transformation operators for an abstract group $g$, we say that the operator $\mcG$ is invariant to $g$ if \vspace{-0.03in}
\begin{equation}
\mcG\circ\mcT_g[\fb]=\mcG[\fb] \text{ ,}\vspace{-0.02in}
\end{equation}
and $\mcG$ is equivariant to $g$ if there exists an equivariant transformation $\mcS_g:\mbU\rightarrow\mbU$, such that
\begin{equation}
\mcG\circ\mcT_g[\fb]=\mcS_g\circ\mcG[\fb] \text{ .}\vspace{-0.06in}
\end{equation}
Considering a mechanical response problem as a practical example in physical systems, we have the input function $\fb(x)$ as the initial location and $\ub(x)$ as the resulting mechanical response in the form of a displacement field. First, let $\mcT_g$ be a translation on the reference frame, i.e., $\mcT_g[\fb]=\tilde{\fb}$, where $\tilde{\fb}(x+g):=\fb(x)$ and $g\in\real^{d}$ is a constant vector. Translational invariance means that translating the input function $\fb$ first and then applying the response operator $\mcG$ will deliver the same result. As such, the resultant physical model does not vary with locations in space, and the Noether's theorem \citep{noether1971invariant} guarantees the conservation of linear momentum. On the other hand, let $\mcT_g$ be a rotation on the reference frame, which rotates the coordinate $x$ as well as the input function, i.e.,  $\mcT_g[\fb]=\tilde{\fb}$, with $\tilde{\fb}(Rx):=R\fb(x)$ and $R$ being an orthogonal matrix. Rotational equivariance means that rotating the input function $\fb$ first and then applying the response operator $\mcG$ will lead to the same result as first applying $\mcG$ and then rotating the output function. %, i.e., $\mcG\circ\mcT_g[\fb](x)=\mcG[\tilde{\fb}](Rx)=\mcT_g\circ\mcG[\fb](x)$. 
As such, the described physical model does not vary under rotations against the origin, and the Noether's theorem \citep{noether1971invariant} guarantees the conservation of angular momentum.

In this work, the proposed INO is designed to handle the following four types of invariance/equivariance:
\begin{enumerate}
\vspace{-0.05in}
\item {\it Translational Invariance.} Translating the reference frame by $g\in\real^{d}$ results in an invariant output, i.e., $\mcG[\tilde{\fb}](x+g)=\mcG[\fb](x)$, where $\tilde{\fb}(x+g):=\fb(x)$.
\vspace{-0.05in}
\item {\it Translational Equivariance.} Translating the reference frame by $g\in\real^{d}$ results in an equivariant translation of the output, i.e., $\mcG[\tilde{\fb}](x+g)=\mcG[\fb](x)+g$, where $\tilde{\fb}(x+g):=\fb(x)$.
\vspace{-0.05in}
\item {\it Rotational Invariance.} Rotating the reference frame results in an invariant output, i.e., for any orthogonal matrix $R\in\real^{d\times d}$, one has $\mcG[\tilde{\fb}](Rx)=\mcG[\fb](x)$, where $\tilde{\fb}(Rx):=R\fb(x)$.
\vspace{-0.05in}
\item {\it Rotational Equivariance.} Rotating the reference frame results in an equivariant rotation of the output,  i.e., for any orthogonal matrix $R\in\real^{d\times d}$, one has $\mcG[\tilde{\fb}](Rx)=R\mcG[\fb](x)$, where $\tilde{\fb}(Rx):=R\fb(x)$.
\end{enumerate}

\vspace{-0.05in}

\subsection{Learning Hidden Physics}

\vspace{-0.05in}

Learning how complex physical systems respond is essential in science and engineering. For decades, physics-based PDEs have been commonly employed to model such systems, and traditional numerical methods \citep{leveque1992numerical} are developed to solve for unobserved system responses. However, the choice of certain governing PDEs is often determined \textit{a priori}, and these PDEs need to be solved numerically for each specified boundary/initial conditions and loading/source terms, which makes classical PDE-based methods insufficient in expressivity and computationally expensive.

Several recent developments in deep NNs have been devoted to providing an efficient surrogate directly from data  \citep{ghaboussi1998autoprogressive,ghaboussi1991knowledge,carleo2019machine,karniadakis2021physics,zhang2018deep,cai2022physics,pfau2020ab,he2021manifold,besnard2006finite}. Among others, neural operators manifest superiority in predicting physical responses as \YY{function mappings}. Contrary to classical NNs that operate between finite-dimensional Euclidean spaces, neural operators are designed to learn mappings between infinite-dimensional function spaces \citep{li2020neural,li2020multipole,li2020fourier,you2022nonlocal,Ong2022,gupta2021multiwaveletbased,lu2019deeponet,lu2021learning,goswami2022physics, gupta2021multiwavelet}. A remarkable advantage of neural operators lies in their resolution independence, which implies that the prediction accuracy is invariant to the resolution of input functions. Moreover, neural operators are generalizable to different input instances, %-- once the network is trained, solving for a new instance of the input function only requires a forward pass, 
and hence they can serve as efficient surrogates in downstream applications. Furthermore, in contrast to classical PDE-based approaches, neural operators can be trained directly from data, and hence requires no domain knowledge nor pre-assumed PDEs. All these advantages make neural operators a promising tool for learning complex physical systems \citep{yin2022simulating,goswami2022physics,yin2022interfacing,you2022physics,li2020neural,li2020multipole,li2020fourier,lu2021comprehensive}.

\YY{Despite the aforementioned advances of neural operators, purely data-driven neural operators still suffer from data challenge. In particular, in order to generalize the solution, they require a large corpus of paired datasets, which is prohibitively expensive in many engineering applications. To resolve this challenge, the physics-informed neural operator (PINO) \cite{li2021physics} and physics-informed DeepONets \cite{goswami2022physics,wang2021learning} are introduced, where a PDE-based loss is added to the training loss as a penalization term. However, these approaches still require \textit{a priori} knowledge of the underlying PDEs, which restricts their applications to (known) PDE-solving tasks. 
%%%%%%%%%%%%%%%%%%%%%%%%%
%Despite the aforementioned, few work has considered neural operators with minimal physical law assumptions such as the translational and rotational invariance/equivariance and their corresponding momentum conservation laws. 
}

\vspace{-5pt}

\subsection{Integral Neural Operators}\label{sec:integralNO}

\vspace{-5pt}
Integral neural operators, first proposed in \cite{li2020neural} and further developed in \cite{li2020multipole,li2020fourier,you2022nonlocal,you2022learning}, have the foundation in the representation of a PDE solution by the Green's function. An integral neural operator is comprised of three building blocks. First, the input function, $\fb(x)\in\mbF$, is lifted to a higher-dimension representation via $\hb(x,0)=\mathcal{P}[\fb](x):=P[x,\fb(x)]^T+p$. Here, $P\in\real^{(d+d_f)\times d_h}$ and $p\in\real^{d_h}$ define an affine pointwise mapping. Next, the feature vector function $\hb(x,0)$ goes through an iterative layer block where the layer update is defined via the sum of a local linear operator, a nonlocal integral kernel operator, and a bias function:  $\hb(\cdot,j+1)=\mathcal{J}_{j+1}[\hb(\cdot,j)]$, for $j=0,\cdots,L-1$. Here, $\hb(\cdot,j)$ is a sequence of functions representing the values of the network at each hidden layer, taking values in $\real^{d_h}$. $\mathcal{J}_1,\cdots,\mathcal{J}_{L}$ are the nonlinear operator layers, which will be further discussed in the later contents. Finally, the output $\ub(\cdot)\in\mbU$ is obtained through a projection layer. A common practice is to project the last hidden layer representation $\hb(\cdot,L)$ onto $\mbU$ as:
$\ub(x)=\mathcal{Q}[\hb(\cdot,L)](x):=Q_2\sigma(Q_1\hb(x,L)+q_1)+q_2$. Here, $Q_1\in\real^{d_{Q}\times d_h}$, $Q_2\in\real^{d_{u}\times d_Q}$, $q_1\in\real^{d_Q}$ and $q_2\in\real^{d_u}$ are the appropriately sized matrices and vectors that are part of the trainable parameter set. $\sigma$ is an activation function, which is often taken to be the popular rectified linear unit (ReLU) function. 

Let $\mcD:=\{(\fb_i,\ub_i)\}_{i=1}^N$ be a support set of observations where the input $\{\fb_i\}\subset\mbF$ is a sequence of independent and identically distributed (i.i.d.) random fields from a known probability distribution $\mu$ on $\mbF$, and $\ub_i(x)\in\mbU$, possibly noisy, is the observed corresponding solution. We aim to learn the system response by building a surrogate operator: \vspace{-0.05in}
$$\tilde{\mcG}[\fb;\theta](x):=\mathcal{Q}\circ\mathcal{J}_{L}\circ\cdots\circ\mathcal{J}_1\circ\mathcal{P}[\fb](x)\approx \ub(x) \text{ ,}\vspace{-0.05in}$$
where the parameter set $\theta$ is obtained by solving the following optimization problem:
\begin{align}
\min_{\theta\in\Theta}\mcL_{\mcD}(\theta)&=\min_{\theta\in\Theta}\mathbb{E}_{\fb\sim\mu}\vertii{\tilde{\mcG}[\fb;\theta]-\mcG[\fb]}\approx \frac{1}{N}\min_{\theta\in\Theta}\sum_{i=1}^N\vertii{\tilde{\mcG}[\fb_i;\theta]-\ub_i} \text{ .}\label{eqn:opt}
\end{align}
%Here $C$ denotes a properly defined cost functional which is often taken as the the mean square error.
The particular choice of an integral neural operator varies by the architecture of the iterative layer block, $\mathcal{J}_{j+1}$. In \cite{li2020neural}, graph neural operators (GNOs) are proposed, where the iterative kernel integration is invariant across layers, i.e., $\mathcal{J}_1=\mathcal{J}_2=\cdots=\mathcal{J}_L:=\mathcal{J}^{GNO}$, with the update of each layer network given by
\begin{align}
\hb(x&,j+1) = \mathcal{J}^{GNO}(\hb(x,j)):=\sigma\left(W\hb(x,j)+\int_\omg \mb(x,y) \hb(x,j)dy + c\right) \text{ ,}\label{eq:gkn_1}\\
\mb(x&,y):= \kappa \left(x,y,\fb(x),\fb(y);v \right)\text{ .}\label{eq:gkn_2}
\end{align}
Here, $W\in\real^{d_h\times d_h}$ and $c\in\real^{d_h}$ are learnable tensors, and $\kappa\in\real^{d_h\times d_h}$ is a tensor kernel function that takes the form of a (usually shallow) NN with learnable parameters $v$. 
%GNOs resemble GNNs in that both follow a message passing framework \citep{brandstetter2021message}, while the discrete aggregation step is substituted by a continuous integral operator. 
\YY{Since the layer update in integral neural operators is formulated as a continuous integral operator, the learned network parameters are resolution-independent: the learned $W$, $c$, and $v$ are close to optimal even when used with different resolutions. 
% However, it was found in \cite{you2022nonlocal} that GNOs may become unstable as the number of network layers grows. To provide a remedy for such instability issues, nonlocal kernel networks (NKNs) were proposed \citep{you2022nonlocal}, in which each iterative layer resembles a fictitious time step in the discrete nonlocal equation, and thus NKNs allow for accelerated learning techniques for deep networks such as the shallow-to-deep technique \citep{haber2018learning}, for which optimal parameters of shallow networks are used as initial guesses of deeper networks. Although NKNs are more stable than GNOs in deep layer limit, one needs to train an additional neural network to account for the reaction term, which is computationally costly. 
Besides GNOs, when both the domain and the discretized points are structured, Fourier Neural Operators (FNOs) \citep{li2020fourier} and Multiwavelet-based Operators (MWT) \citep{gupta2021multiwaveletbased} can be employed. In FNOs, the fast Fourier transform is employed to evaluate the integrals, which presents superior efficiency. 
Nevertheless, despite the rapid advancement in neural operators, existing methods fail to preserve the invariance/equivariance properties under translation and rotation operations.}

\subsection{Invariant and Equivariant Neural Networks}

\vspace{-0.05in}

Recently, invariant and equivariant NNs have been developed in the context of convolutional neural networks \citep{lang2020wigner,chirikjian2000engineering,knapp2001representation} and graph neural networks (GNNs) \citep{bruna2013spectral,defferrard2016convolutional,kipf2016semi}, and their effectiveness is demonstrated via a variety of machine learning tasks, such as in image classification \citep{cohen2016group,cohen2016steerable,weiler2019general,romero2020group} and dynamical system modelling \citep{rezende2019equivariant,satorras2021n}. To achieve equivariance, the authors in \cite{thomas2018tensor,fuchs2020se} proposed to utilize spherical harmonics to compute a set of basis for transformations between higher-order representations. As another line of efforts \citep{schutt2017quantum,klicpera2020directional,anderson2019cormorant,miller2020relevance,satorras2021n}, GNNs were considered based on a message passing framework \cite{brandstetter2022message}, in which the translational and rotational invariance/equivariance were imposed by specially designed edge and node update operations. However, the above-mentioned invariant/equivariant networks are restricted to a discrete ``vector-to-vector'' mapping, and the learned parameters cannot be reused in networks of different input/output resolutions, \YY{which hinders their application to learn hidden physics laws in the form of function mappings}. \YY{Therefore, the goal in this work is to design neural operators and impose minimal physical law assumptions as the translational and rotational invariance/equivariance, so as to  provide a data-driven model form that learns complex physical systems with guaranteed momentum conservation.} 

\vspace{-5pt}

\section{Neural Operators with Momentum Conservation}

\vspace{-5pt}

In this section, we develop the invariant/equivariant architecture based on integral neural operators. 
%First, in Section \ref{sec:integralNO} we review integral neural operators as our baseline. In Section \ref{sec:ino} we present the proposed INO architecture for scalar- and vector-valued function problems, together with theoretical results for their invariance properties.
Our developments have two-folds. First, a node embedding update scheme is proposed, that is physically invariant and preserves invariance/equivariance to translations and rotations on a continuous domain. The essential idea is to devise a message passing neural network where its arguments and relevant representation embeddings are invariant to transformations. As such, we can convert transformation-sensitive representations to their transformation-invariant counterparts. Second, to handle general domains and accelerate training, we also modify the GNO architecture in Eqs.~\eqref{eq:gkn_1}-\eqref{eq:gkn_2} in such a way that each layer resembles a discretized time-dependent nonlocal equation \cite{you2022nonlocal}. As a result, our proposed architecture can be seen as a resemblance with translational and rotational invariant/equivariant nonlocal differential equations, allowing for generalization of its optimal parameters from shallow to deep networks.

To establish a transformation-invariant kernel in Eq.~\eqref{eq:gkn_2}, we introduce two types of transformation-invariant quantities as arguments to the kernel function: the vector Euclidean norm of the edge between $x$ and $y$, i.e., $\verti{y-x}$, and the orientation of the vector $y-x$ with respect to a local reference vector in the undeformed coordinates. For example, the local reference edge in a rectangular domain can be either the horizontal or vertical edge of the rectangle. In the perspective of numerical implementation, one can take the vector formed by any two fixed nodes as the reference edge, as illustrated in Figure \ref{fig:equivariance}(b). In physical problems with 2D domains, $\omg\subset\real^2$, we pass in as kernel arguments the decomposed Euclidean norm in the following form:
\begin{equation}\label{eq:norm_decompose}
\overline{y-x} := [\verti{y-x} \cos\theta,
          \verti{y-x} \sin\theta] \text{ ,}
\end{equation}
where $x$ and $y$ are the source and target nodes connected by the edge, and $\theta$ denotes the computed local orientation. Similarly, for $\omg\subset\real^3$, three kernel arguments are passed in, based on two angles from the computed local orientation. Here, we point out that the idea of parameterizing the edge feature with its Euclidean norm, $\verti{y-x}$, was also employed in the equivariant graph neural network proposed in \cite{satorras2021n}. However, our approach has for the first time considered the local edge orientation together with its Euclidean norm, which makes the resulting network more expressive. As will be demonstrated in the ablation study in Section \ref{sec:exp}, an Euclidean norm-based kernel architecture may not be sufficiently expressive. %comparing with our baseline neural operators. % for problems that involve solving PDEs.

%Once the Euclidean norm and the local orientation of the edge become available, we pass in as kernel arguments the decomposed Euclidean norm based on the edge orientation in the following form:
% \begin{equation}\label{eq:norm_decompose}
% \overline{y-x} := [\verti{y-x} \cos\theta,
%           \verti{y-x} \sin\theta]^T,
% \end{equation}
% where $\xb_i$ and $\xb_j$ are the source and target nodes connected by the edge, and $\theta$ denotes the computed local orientation.

% Thus, we introduce an additional transformation-invariant quantity as the argument to the kernel function: the orientation of an edge with respect to a local reference edge. For example, the local reference edge in a rectangular domain can be either the horizontal or vertical edge of the square. In the perspective of numerical implementation, we can take the vector formed by any two fixed nodes as the reference edge. Once the Euclidean norm and the local orientation of the edge become available, we pass in as kernel arguments the decomposed Euclidean norm based on the edge orientation in the following form:
% \begin{equation}\label{eq:norm_decompose}
% \xb_i-\xb_j = \begin{bmatrix}
%           \vertii{\xb_i-\xb_j} \cos\theta \\
%           \vertii{\xb_i-\xb_j} \sin\theta
%          \end{bmatrix} \text{ ,}
% \end{equation}
% where $\xb_i$ and $\xb_j$ are the source and target nodes connected by the edge, and $\theta$ denotes the computed local orientation.

\textbf{INO for scalar-valued functions.} We first consider the scenario where the output function takes scalar values, i.e., $d_u=1$, and the physical system is both translation- and rotation-invariant. Examples in this category involve the prediction of energies in environmentally-powered systems \citep{cammarano2012pro}, pressure monitoring in subsurface flows \citep{fumagalli2011numerical}, and the prediction of damage field in brittle fracture \citep{fan2022meshfree}. In this context, we propose the the following INO-scalar architecture: for the lifting block, we only pass in the Euclidean norm of the input function:
%\vspace{-5pt}
\begin{equation}\label{eq:inos_p}
\hb(x,0)=\mathcal{P}[\fb](x):=P\verti{\fb(x)}+p \text{ ,}%\vspace{-5pt}
\end{equation}
where $P,p\in\real^{d_h}$. Then, for the iterative layer, we introduce a fictitious time step, $\tau$, and regard different layer features as the solution of a time-dependent nonlocal equation at different time instances:
\begin{align}
&\hb(x,(j+1)\tau):=\hb(x,j\tau)+\tau\sigma\left(W\hb(x,j\tau)+\int_\omg \mb(x,y)\hb(y,j\tau) dy + c\right) \text{ ,}\label{eq:inos_1}\\
&\mb(x,y):= \kappa \left(\overline{y-x},\verti{\fb(x)},\verti{\fb(y)};v \right)\text{ .}\label{eq:inos_2}
\end{align}
Finally, the projection block is taken as a 2-layer multilayer perceptron (MLP), same as the other integral neural operators. Herein, we notice that the architecture in Eq.~\eqref{eq:inos_1} resembles the time-dependent nonlocal equation: if we divide both sides of Eq.~\eqref{eq:inos_1} by the fictitious time step, $\tau$, the term $(\hb(x,(j+1)\tau)-\hb(x,j\tau))/\tau$ corresponds to the discretization of a first order derivative so that this architecture can indeed be interpreted as a nonlinear differential equation in the limit of deep layers, as $\tau\to 0$. Hence, in practice we can employ the shallow-to-deep learning technique \citep{haber2018learning,you2022nonlocal}, that corresponds to training the network for increasing values of network layers and using optimal parameters obtained with $L$ layers as initial guesses for the $\tilde{L}$-layer INO. Here $\tilde{L}>L$. Moreover, we point out that all network arguments are translational and rotational invariant, and hence we have the following theorem, with its detailed proof provided in \ref{app:a}:
\begin{theorem}[Invariance for INO-scalar] \label{thm:INO-scalar}
The INO-scalar architecture proposed in Eqs.~\eqref{eq:inos_p}-\eqref{eq:inos_2} is translational and rotational invariant. That means, when translating the reference frame by $g\in\real^d$ and rotating it by an orthogonal matrix $R\in\real^{d\times d}$, the following property holds true:
$$\tilde{\mcG}[\tilde{\fb};\theta](Rx+g)=\tilde{\mcG}[\fb;\theta](x) \text{ ,}$$
where $\tilde{\fb}(Rx+g):=R\fb(x)$.
\end{theorem}

\begin{figure}[!t]\centering
     \includegraphics[width=.6\linewidth]{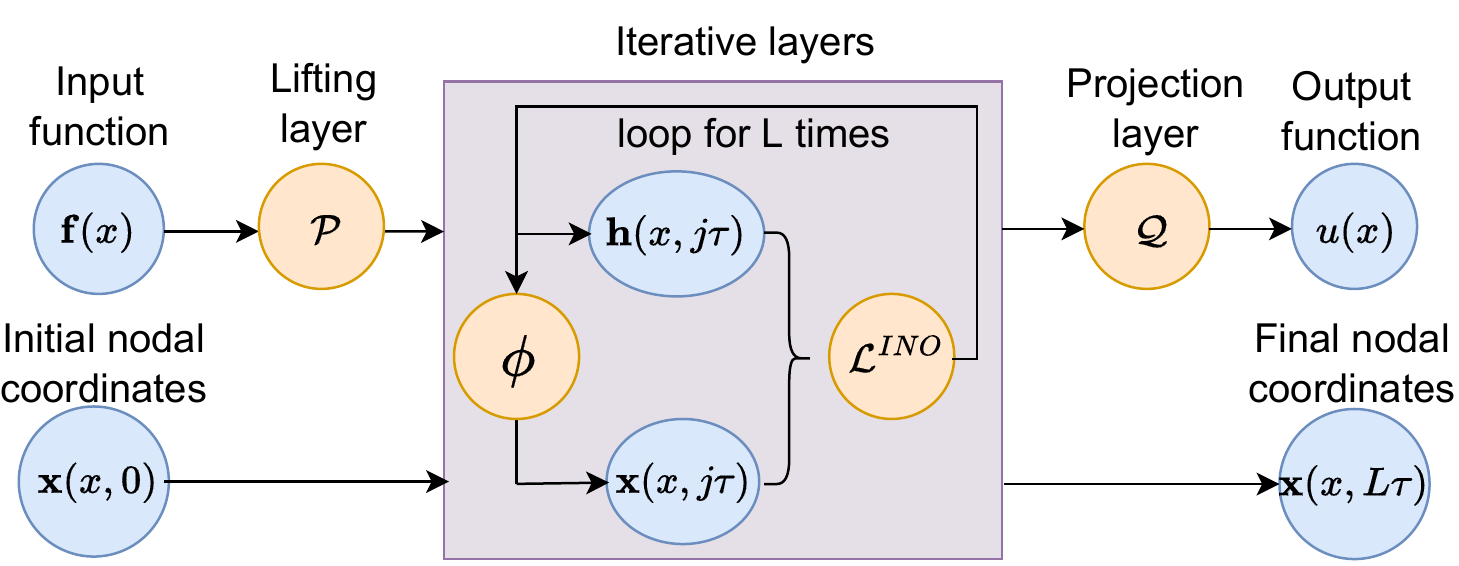}
     \caption{An illustration of the proposed INO architecture. We start from input $\fb(x)$ and the initial coordinate $\xb(x,0):=x$. Then, the iterative layers are built as integral operators based on invariant kernel functions, to obtain a frame-invariant layer feature vector update $\hb(x,j)$ and a frame-dependent coordinate update $\xb(x,j)$ which embeds the coordinate rotation information. Lastly, we project the last hidden layer representation (for scalar-valued functions) or the coordinate update (for vector-valued functions) to the target function space. \vspace{-0.1in}}
     \label{fig:ino_architecture}
\end{figure}

\textbf{INO for vector-valued functions.} We now further consider the case where the output function takes vector values, and hence the output should rotate equivariantly with the rotation of the reference frame. Under this scenario, rotation-equivariant property is required to achieve the conservation of angular momentum. As such, besides the layer feature function $\hb(x,j\tau)$, $j=0,\cdots,L$, we further introduce an additional coordinate function, $\xb(x,j\tau)$, which is defined on domain $\omg$ and takes values in $\real^d$. Then, the key is to preserve invariance to rotations on $\hb$, as well as equivariance on $\xb$. In light of this, when translational invariance and rotational equivariance are desired, we propose the following INO-vector architecture (cf. Figure~\ref{fig:ino_architecture}): for the lifting block, we provide the Euclidean norm of the original feature to $\hb$ and carry the coordinate information into $\xb$:
\begin{align}
&\hb(x,0)=\mathcal{P}[\fb](x):=P\verti{\fb(x)}+p \text{ ,}\label{eq:inov_p1}\\
&\xb(x,0):=x\label{eq:inov_p2} \text{ ,}
\end{align}
with $P,p\in\real^{d_h}$. Then, the $(l+1)$-th iterative layer network update of a $L$-layer INO is defined as,
\begin{align}
&\hb(x,(j+1)\tau):=\hb(x,j\tau)+\tau\sigma\left(W\hb(x,j\tau)+\int_\omg \mb(x,y)\hb(y,j\tau) dy + c\right),\label{eq:inov_1}\\
&\xb(x,(j+1)\tau):=\xb(x,j\tau)+\tau\int_\omg (x-y)\phi(\mb(x,y)\hb(y,j\tau);w) dy\text{ ,}\label{eq:inov_2}\\
&\mb(x,y):= \kappa \left(\overline{y-x},\verti{\fb(x)},\verti{\fb(y)};v \right)\text{ .}\label{eq:inov_3}
\end{align}
% where the message function $\mb_{ij}$ and the coordinate embeddings $\xb^{l+1}_i$ are respectively given as,
% \begin{align}\label{eq:msg_En}
% \mb_{ij}:=& \phi_{e}(\xb_i-\xb_j,b(\xb_i),b(\xb_j);v)\hb^l_j\text{ ,}\\
% \label{eq:coords_En}
% \xb^{l+1}_i:=&\xb^{l}_i+C\Delta t\int_\omg (\xb^{l}_i-\xb^{l}_j)\phi_x(\mb_{ij}; \wb) d\xb_j\text{ ,}
% \end{align}
Finally, we define the projection block and the output function as,
\begin{equation}\label{eq:inov_q}
\ub(x)=\mcQ[\xb(x,L\tau)](x):=\xb(x,L\tau)-x \text{ .}    
\end{equation}
Here, $\kappa$ and $\phi$ are two separate (usually shallow) MLPs for computing edge and coordinate messages, with $v$ and $w$ being the corresponding trainable parameters, respectively. In Eq.~\eqref{eq:inov_2}, $\phi$ takes as input the edge embeddings and outputs a scalar representing the weight associated with its neighboring node. The nodal positions are then updated as the weighted sum of coordinate differences. When considering the output function $\ub$ as the displacement field and $\xb$ as the updated position of material points, the proposed INO-vector architecture can be seen as an analogue to the particle-based methods, since both approaches describe the motion of a particle by the summation of forces due to its neighboring particles. Additionally, the INO architecture preserves the continuous integral treatment of the interactions between nodes that characterizes neural operators. As a result, INO also permits resolution independence with respect to the inputs, and hence serves as a surrogate operator between function spaces. Formally, we have the following theorem, with the detailed proof provided in \ref{app:a}:
\begin{theorem}[Invariance/equivariance for INO-vector] \label{thm:INO-vector-1}
The INO-vector architecture proposed in Eqs.~\eqref{eq:inov_p1}-\eqref{eq:inov_q} is translation-invariant and rotation-equivariant. That means, when translating the reference frame by $g\in\real^d$ and rotating it by an orthogonal matrix $R\in\real^{d\times d}$, the following property holds true:
$$\tilde{\mcG}[\tilde{\fb};\theta](Rx+g)=R\tilde{\mcG}[\fb;\theta](x) \text{ ,}$$
where $\tilde{\fb}(Rx+g):=R\fb(x)$.
\end{theorem}
%Besides the material displacement tracking discussed in Section \ref{sec:back}, other exemplar physical problems with translational invariance and rotational equivariance include computational fluid dynamics where the fluid velocity is of interest \citep{anderson1995computational}, external loading estimation in robot manipulation \citep{colome2013external}, etc. 

%Analogously, a translation- and rotation-equivariant INO-vector architecture can be defined. We leave further discussion in the Appendix.

%, with $\verti{\omg}$ chosen as the area of $\omg$, which can be seem as a reciprocal of the number of interacting nodes to mitigate the spatial variability of edge features. 

% In dynamical systems where the velocity $\vb$ is a variable of interest, Eq.~\eqref{eq:coords_En} can be replaced by,
% \begin{align}\label{eq:vel_En}
% \vb^{l+1}_i:=&\vb^{l}_i+\Delta t\phi_{v}(\hb^l_i)\vb^{l}_i \nonumber\\&~~~~+C\Delta t\int_\omg (\xb^{l}_i-\xb^{l}_j)\phi_x(\mb_{ij}; \wb) d\xb_j\text{ ,}\\
% \xb^{l+1}_i:=&\xb^{l}_i+\Delta t\vb^{l+1}_i\text{ .}
% \end{align}

% Note that Eq.~\eqref{eq:coords_En} is not needed for scalar functions. In this case, the proposed INO architecture becomes similar to that of a NKN. In addition to the translational and rotational invariance, a major improvement of INO lies in the fact that only one NN needs to be learned, which boosts computational efficiency.

%\YY{[add theorems for rotational invariant and conservation laws]}

\vspace{-0.1in}

\section{Experiments}\label{sec:exp}

\vspace{-0.1in}

\begin{figure}[!t]\centering
    \includegraphics[width=0.75\linewidth]{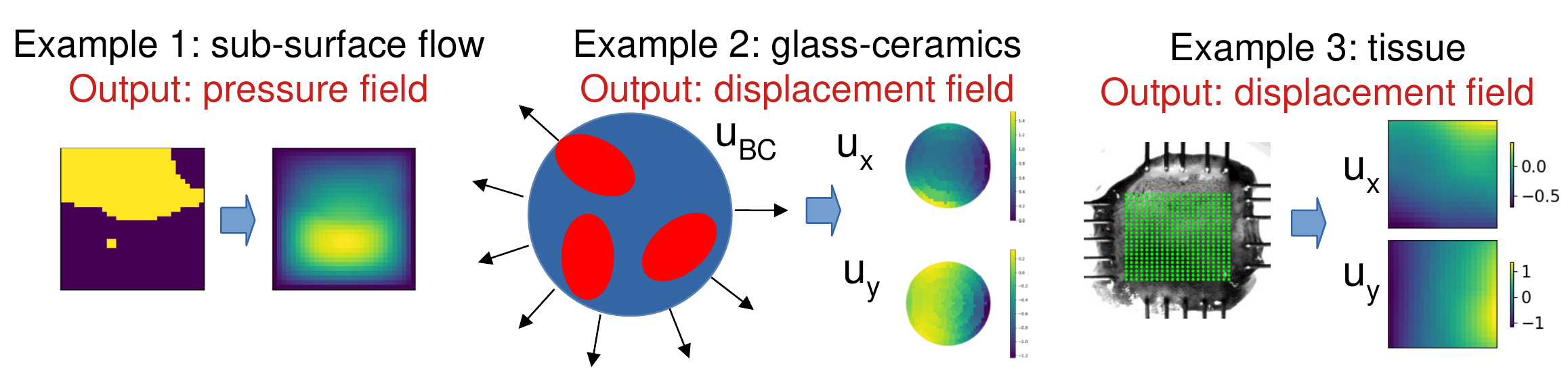}
    \caption{Illustration of input/output settings in our three exemplar problems.\vspace{-0.1in}}
    \label{fig:experiments}
\end{figure}

In this section, we demonstrate the empirical effectiveness of INOs. Specifically, we conduct experiments on both 2D synthetic and real-world datasets% (cf. Figure~\ref{fig:experiments})
, and compare the proposed INO against three data-based neural operator baselines -- GNOs \cite{li2020neural}, FNOs \cite{li2020fourier}, and MWTs \cite{gupta2021multiwavelet}, a physics-informed neural operator baseline (PINO \cite{li2021physics}), and an equivariant GNN (EGNN \cite{satorras2021n}). Herein, we employ $L=4$ iterative layers for all neural operators. All experiments are tested using PyTorch with Adam optimizer. For fair comparison, we tune the hyperparameters for each method, including the learning rates, decay rates, and regularization coefficient, to minimize the error on the validation dataset. In all tests, we report the averaged relative error, $||\ub_{i,pred}-\ub_{i}||/||\ub_{i}||$, as the comparison metric (lower means better). An illustration of our three test examples are provided in Figure \ref{fig:experiments}, with further details of each dataset and experimental settings provided in \ref{app:b}.

\vspace{-0.05in}

\subsection{Synthetic dataset: sub-surface flow}

\vspace{-0.05in}

As a benchmark for scalar-valued output functions, we consider the modeling of 2D sub-surface flow through a porous medium with heterogeneous permeability field.
%, and compare the results obtained using GNO \citep{li2020neural,li2020multipole}, FNO \citep{li2020fourier}, PINO \citep{li2021physics}, MWT \citep{gupta2021multiwavelet}, and the proposed INO-scalar. 
The high-fidelity synthetic simulation data in this example is described by the Darcy flow, which has been considered in a series of neural operator studies \citep{li2020neural,li2020multipole,li2020fourier,lu2021comprehensive,you2022nonlocal,you2022learning}. Specifically, the governing differential equation is defined as:
\begin{equation}\label{eq:darcy_PDE}
\begin{split}
    -\nabla \cdot (\fb(x)\nabla \ub(x)) &= 1,\, \forall \;\; x\in\omg:=[0,1]^2,\\
    \text{subjected to}\;\;\;\ub_{BC}(x) &= 0, \, \forall \;\; x \in \partial \Omega,
\end{split}
\end{equation}
where $\fb$ is the conductivity field, and $\ub$ is the hydraulic head (both are scalar-valued functions). In this context, we aim to learn a solution operator of the Darcy's equation that maps each realization of the conductivity field $\fb(x)$ to the corresponding solution field $\ub(x)$. For training, we employ the dataset from \cite{li2020neural}, where the conductivity field $\fb(x)$ is modeled as a two-valued piecewise constant function with random geometries such that the two values have a ratio of $4$. Then, the ground-truth solutions of $\ub(x)$ were generated using a second-order finite difference scheme on a fine resolution of $241\times 241$ grids and downsampled to $16\times 16$ and $31\times 31$. Herein, we split the provided test samples into $40$ samples as validation for hyperparameter tuning and $40$ samples as test. To investigate the performance of each model under small data regime, we train with $N^{\text{train}}=\{5,10,20,40,100\}$ numbers of labelled data pairs. The dimension of representation is set to $d_h=64$, and the kernel $\kappa$ is modeled by a 3-layer MLP with width $(n, 512, 1024, d_h^2)$, where $n$ is the number of kernel arguments for each architecture. For 2D problems, $n=6$ for GNOs and $n=4$ for INOs.

% \begin{figure}[!t]\centering
%     \includegraphics[width=.99\linewidth]{figures/lps_tissue_loss.pdf}
%     \caption{An illustration of the proposed INO architecture.}
%     \label{fig:darcy_loss}
% \end{figure}

% \begin{table}[]
%     \centering
%     \begin{tabular}{|c|c|c|c|c|c|}
%     \hline 
%         \multirow{2}{*}{Model} & \multicolumn{3}{c|}{Peridynamics} & \multicolumn{2}{c|}{Tissue}\\
%         \cline{2-6}
%         & 10 & 40 & 100 & 10 & 100 \\
%         \hline 
%          GNO, train &  9.482\% &  4.089\%  & 1.584\% &  9.632\%  & 6.761\%\\
%          GNO, test &  31.252\% &  10.163\%  & 8.495\% & 38.356\%  & 14.150\%\\
%          INO, train &  4.430\% &  7.202\%  & 7.278\% &  4.501\%  & 3.645\%\\
%          INO, test &  \textbf{12.652}\% &  \textbf{8.188}\%  & \textbf{7.944}\% &  \textbf{17.371}\%  & \textbf{6.376}\%\\
%          \hline 
%     \end{tabular}
%     \caption{Peridynamics and tissue data training and test loss. The 2nd row indicate the number of training samples. The learned peridynamics model is tested with 40 samples, whereas the learned tissue model is tested with 4500 samples.}
%     \label{tab:lps_tissue}
% \end{table}

\begin{figure*}[h!]
\centering%
\includegraphics[width=0.31\columnwidth]{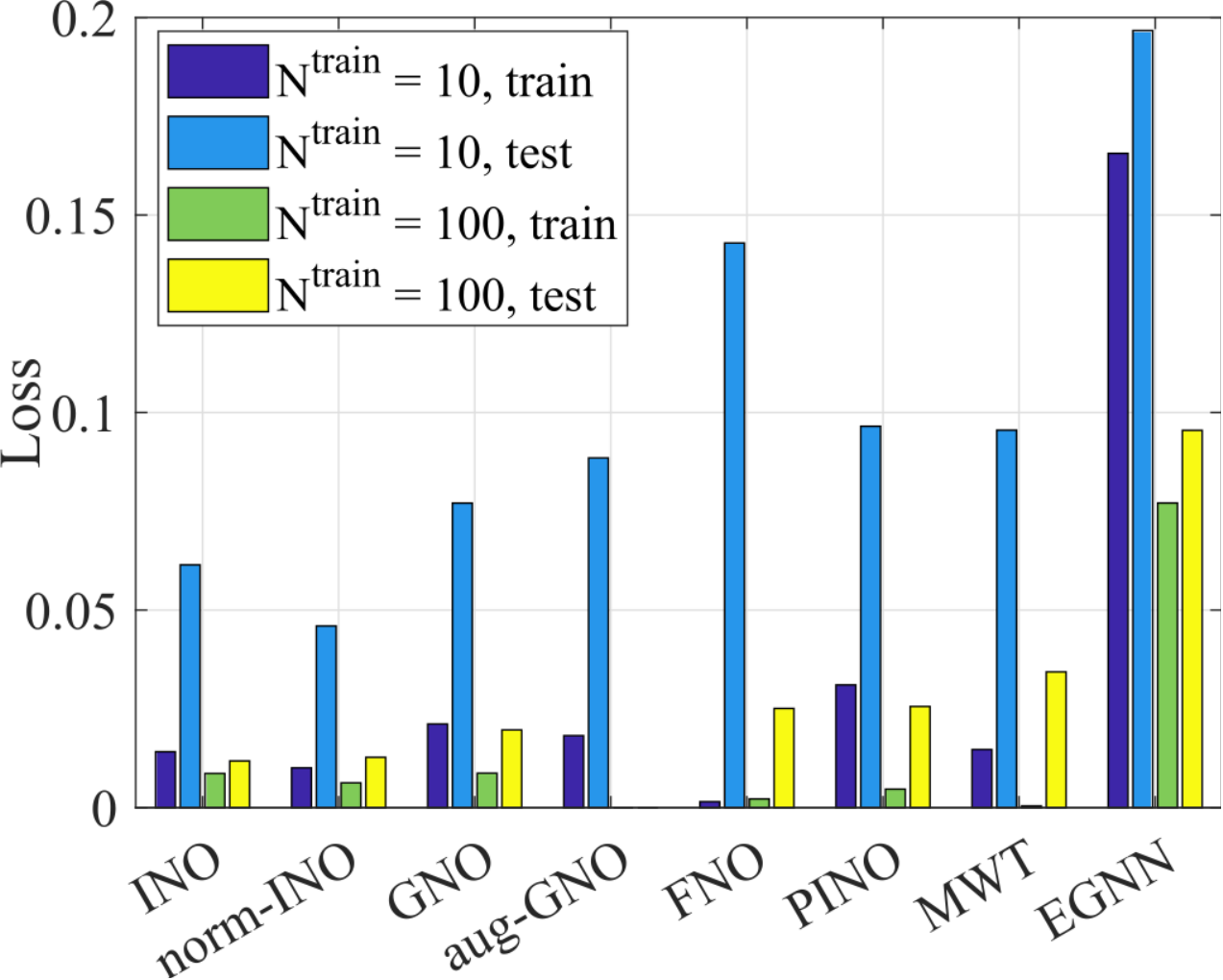}
\includegraphics[width=0.31\columnwidth]{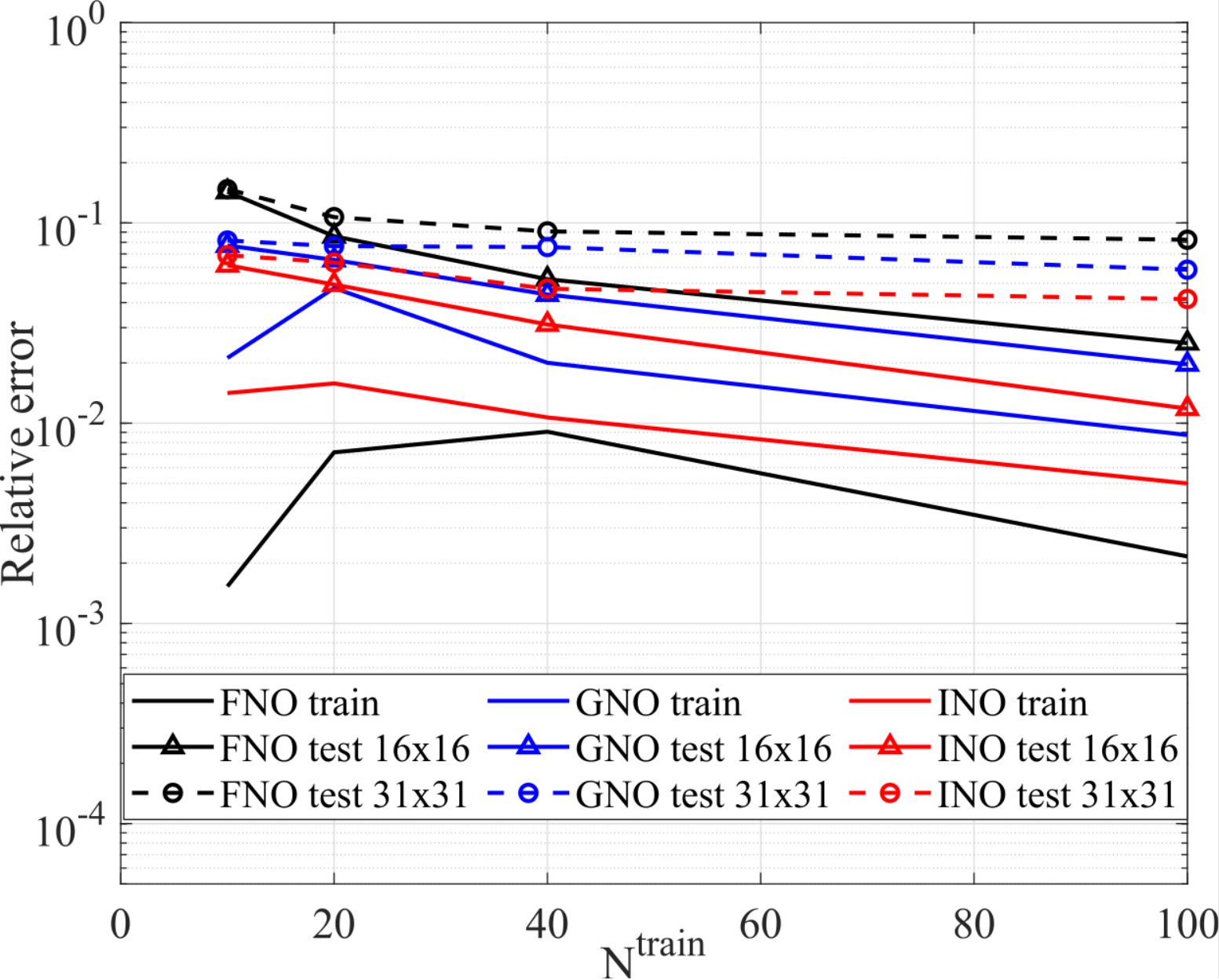}
\includegraphics[width=0.36\columnwidth]{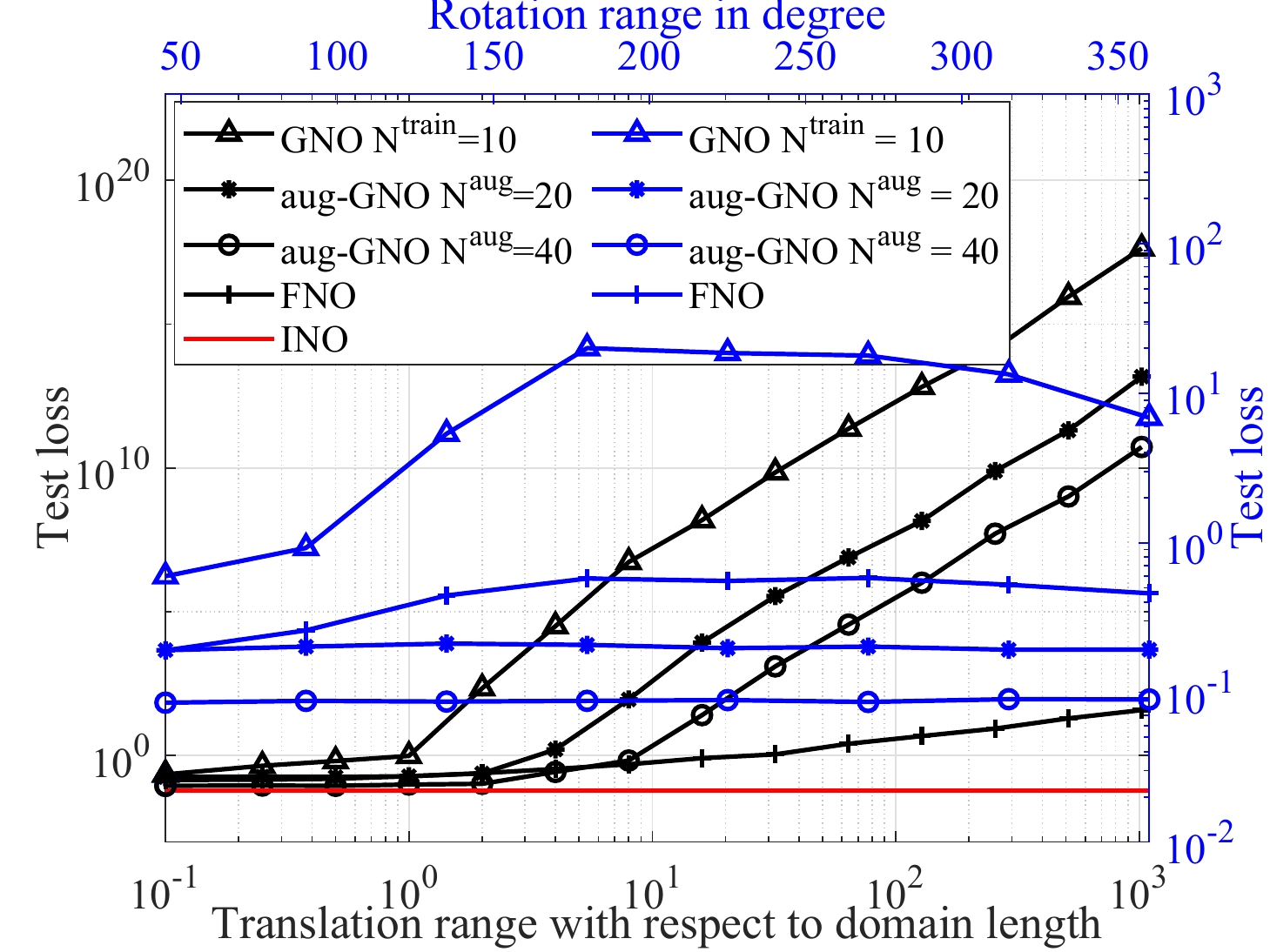}
    \caption{Results for cases with scalar-valued output function. Left: comparison between INO, GNO, norm-INO, aug-GNO, and other baseline models in small and medium data regimes. Middle: comparison between FNO, GNO, and INO with varying numbers of training samples and test on grids with different resolutions. Right: Generalization results to test samples with translated and rotated reference frames.}
    \label{fig:darcy}
\end{figure*}

\textbf{Ablation study. }We first conduct an ablation study on the proposed algorithm, with four settings: 1) The original INO-scalar architecture; 2) The original GNO architecture; With this test, we aim to investigate the expressivity of our invariant kernel compared to the original kernel of \eqref{eq:gkn_2}. 3) The INO-scalar architecture, with its kernel $\kappa$ depending on the Euclidean norm of the edge only (denoted as ``norm-INO''); With this test, we study if the local edge orientation, $\theta$, plays an important role in the model. 4) The GNO architecture with training data augmentation (denoted as ``aug-GNO''), where we train the GNO model with additional translated/rotated samples.  Specifically, we translate the reference frame by setting $\tilde{\omg}\in R([C_x,1+C_x]\times [C_y,1+C_y])$. Herein, the randomly generated constants $C_x,C_y\sim \mcU[-1,1]$, where $\mcU[-1,1]$ denotes the uniform distribution on $[-1,1]$. Similarly, for rotation we randomly generate $C_\theta\sim\mcU[0,2\pi]$ and rotate the reference coordinates counter-clockwisely. For each training sample, we repeat the above process to augment the training set for 3 times. With this test, we investigate if our invariant architecture outperforms an invariance-agnostic approach with data augmentation. On each of these four settings, we report the training and test errors with $N^{\text{train}}=\{10,100\}$\footnote{Since the GNO model could not finish its training in the ``aug-GNO, $N^{\text{train}}=100$'' case within 72 hours, we only report the results from $N^{\text{train}}=10$ for this setting.} to study the performance of each model in small and medium data regimes. Unless otherwise stated, all trainings and tests are performed on $16\times 16$ structured grids. We plot the results in the top left of Figure \ref{fig:darcy}, with further error comparisons provided in Table \ref{tab:darcy_results} of \ref{app:b}.

%%%%%%%%%%%%%%%%%%%%%%

As shown in the top left of Figure \ref{fig:darcy}, INOs outperform GNOs in both small and medium data regimes.  When $N^{\text{train}}=100$, INOs present $1.2\%$ test error while GNOs have $2.0\%$ test error. When $N^{\text{train}}=10$, INOs have $6.1\%$ test error, which also outperforms GNOs (with $7.7\%$ test error) by $20\%$. This is possibly due to the fact that INOs have only 4 arguments in its kernel function $\kappa$, while GNOs have 6. Hence, INOs are less likely to overfit with small and medium data. Surprisingly, the data-augmentation strategy did not help much, and the aug-GNOs show a similar accuracy to the original GNOs. On the other hand, when comparing the performance between INOs and norm-INOs, we notice that norm-INOs achieves the best performance in the small data regime, possibly due to the fact that it further reduces the number of kernel arguments to 3. However, when we increase $N^{\text{train}}$ to $100$, the performance of norm-INOs deteriorates due to the lack of flexibility. Hence, in this paper we focus on INOs with a 4-argument kernel, since it outperforms GNOs in both small and medium data regimes, and has better expressivity than the norm-INOs in the medium data regime. 

\textbf{Comparison with more baselines. }
We now present the comparison with other baselines methods by comparing the test errors of INO-scalar with GNO, FNO, MWT, PINO, and EGNN. To obtain a fair comparison, the numbers of trainable parameters for all models are within the range of $[4.2M,4.8M]$ (see Table \ref{tab:darcy_para} in \ref{app:b} for further details). As shown in the top left of Figure \ref{fig:darcy} and Table \ref{tab:darcy_results}, our proposed INOs have achieved the best accuracy in both small ($N^{\text{train}}=10$) and medium ($N^{\text{train}}=100$) data regimes. Here, we note that in the PINO architecture, we follow \cite{li2021physics} and add the error associated with the PDE governing equation as a penalization term in the loss. The penalty parameter was tuned together with other hyperparameters, so as to optimize the validation error. With this test, we aim to compare the efficacy of INOs with another popular physics-informed approach, where full physics knowledge is infused via a soft constraint (when the governing equation is known). As can be seen from the results, INO still outperforms PINO even in the small data regime, where the best PINO presents $9.7\%$ error. Generally, when the PDE loss has a large weight, PINO gets closer to PINN and suffers from slow convergence. On the other hand, our INOs embed the physics knowledge in the architecture instead of a soft constraint, and do not suffer from the challenges in optimization.

\begin{table}[]
    \centering
    {\small\begin{tabular}{|c|ccc|cc|}
    \hline 
        {Dataset} & \multicolumn{3}{c|}{Glass-Ceramics} & \multicolumn{2}{c|}{Biological Tissue}\\
        \cline{1-6}
        $N^{\text{train}}$& 10 & 40 & 100 & 10 & 100 \\
        \hline 
         GNO, train &  9.48\% &  4.09\%  & 1.58\% &  9.63\%  & 6.76\%\\
         GNO, test &  31.25\% &  10.16\%  & 8.50\% & 38.36\%  & 14.15\%\\
\hline         
         INO, train &  4.43\% &  7.20\%  & 7.28\% &  4.50\%  & 3.65\%\\
         INO, test &  \textbf{12.65}\% &  \textbf{8.19}\%  & \textbf{7.94}\% &  \textbf{17.37}\%  & \textbf{6.38}\%\\
\hline                 
         FNO, train & - & - & - & 8.49\%  & 3.95\%\\
         FNO, test & - & - & - & 36.73\%  & 8.49\%\\
\hline                  
         MWT, train & - & - & - & 21.16\%  & 2.24\%\\
         MWT, test & - & - & - & 41.79\%  & 7.57\%\\
         \hline 
    \end{tabular}}
\caption{Results for cases with vector-valued output function. Training and test errors in small and medium data regimes, where bold numbers highlight the best method for each case.}
    \label{fig:vector_table}
\end{table}

\begin{figure*}[h!]
\centering%
\includegraphics[width=0.6\columnwidth]{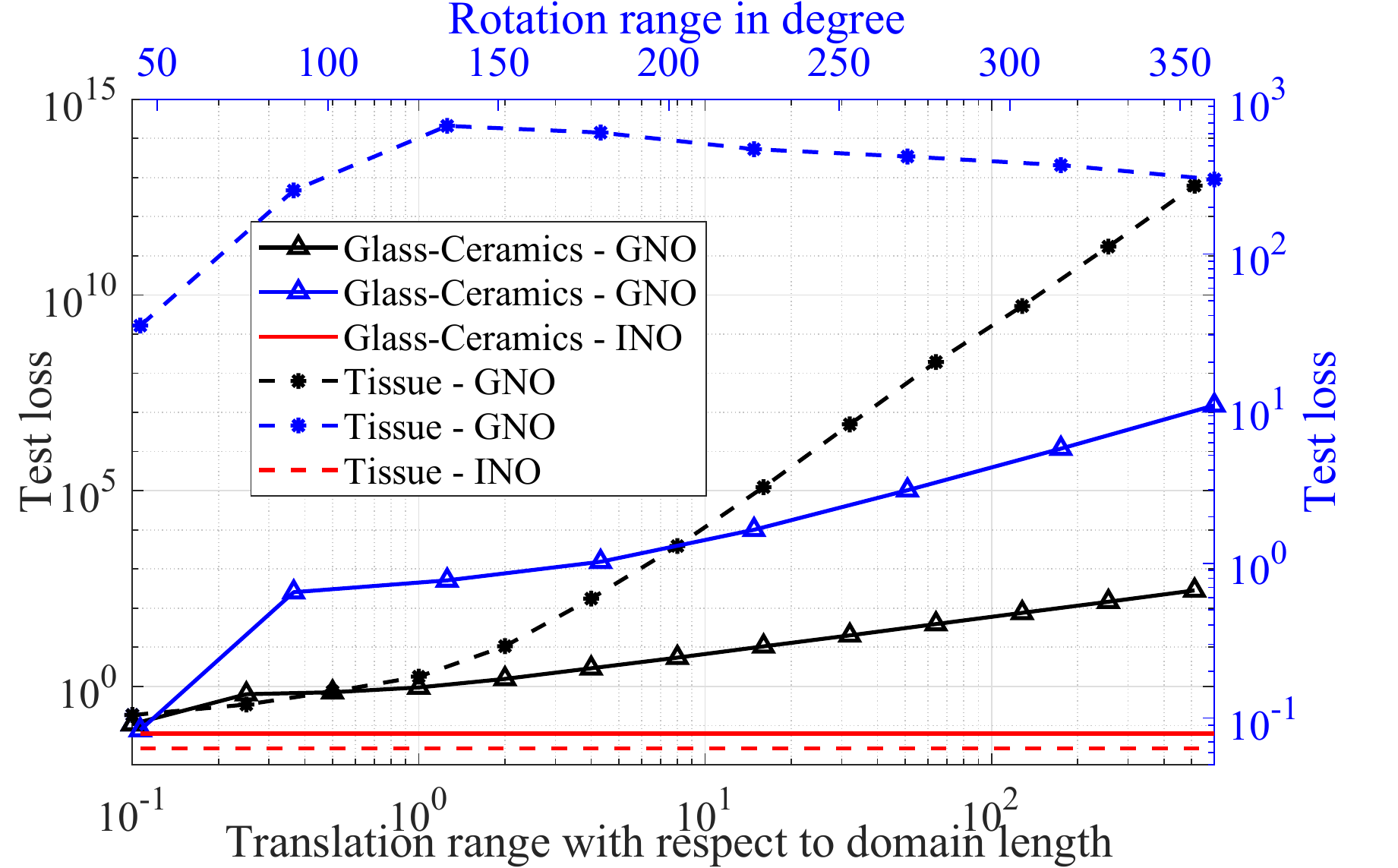}
\caption{Generalization results to test samples with translated and rotated reference frames, for cases with vector-valued output function.}
    \label{fig:vector}
\end{figure*}

\textbf{Performance with the change of $N^{\text{train}}$ and resolutions. } Next, we further compare our INOs with two popular baseline neural operators, FNOs and GNOs, for $N^{\text{train}}=\{5,10,20,40,100\}$. For the purpose of testing generalization properties with respect to resolution, we train all models on $16\times 16$ structured grids, and consider two testing datasets: the ``original-resolution'' dataset with $16\times 16$ grids, and a ``fine-resolution'' dataset with $31\times 31$ grids. In the top right plot of Figure \ref{fig:darcy}, we report the training and test errors on both original- and fine-resolution datasets for each model, with further visual comparison on an exemplar test instance provided in Figure \ref{fig:plot_cross_resolution} of \ref{app:b}. One can observe that INOs consistently out-performe GNOs and FNOs. Among all three neural operators, in the small data regime FNOs have the smallest training error and the largest test error, indicating that they suffer more from overfitting. This observation is also consistent with the findings in \cite{you2022nonlocal,you2022learning}. When comparing the test results across two resolutions, we observe that test errors at different resolutions remain on a similar scale for all three methods. This fact again verifies the capability of neural operators in handling different resolutions. On the contrary, the EGNN model learns a vector-vector mapping rather than a function-function mapping, which makes it not resolution-invariant: testing EGNN on $31\times 31$ grid samples yields $492\%$ error.

%\textbf{Resolution independence tests. }

\textbf{Translation and rotation tests. }Lastly, we verify the capability of INOs in handling translated and rotated samples. In the translated test dataset, we translate the reference frame of each test sample by shifting it onto a new domain, $\tilde{\omg}\in[C_x,1+C_x]\times [C_y,1+C_y]$. Here the movement of domain is randomly generated as  $C_x,C_y\sim \mcU[-C,C]$ and the constant $C$ defines the translation range. Similarly, in the rotated dataset we randomly generate $C_\theta\sim[0,C]$ for each test sample, and rotate the reference frame by $C_\theta$ counter-clockwisely. The test errors of GNOs, FNOs, and INOs are reported in the bottom plot of Figure \ref{fig:darcy}, along with results from GNOs with data-augmentation, where $N^{\text{aug}}$ represents the total number of augmented training samples in addition to $N^{\text{train}}$. Perhaps unsurprisingly, while INOs exhibit invariant performance on translated and rotated datasets, the performances of GNOs and FNOs deteriorate as the range of translation/rotation, $C$, increases. When comparing the results from original GNOs and aug-GNOs, we notice that the data-augmentation trick is generally helpful in handling transformations, although it increases the training dataset size and requires longer training time. In Figure \ref{fig:plot_trans_rot_pred} of \ref{app:b} we plot the test results of INO, GNO, and FNO on an exemplar instance with translated and rotated frame coordinates, respectively, where one can see that the INO solutions are invariant while the solutions from GNO and FNO are not.

\vspace{-10pt}
\subsection{Synthetic dataset: glass-ceramics deformation}

\vspace{-0.05in}

In this example, we study material deformation in a glass-ceramic specimen as a prototypical exemplar on the heterogeneous material response prediction. A glass-ceramic material is the product of controlled crystallization of a specialized glass composition, which results in a microstructure of one or more crystalline phases within the residual amorphous glass \citep{prakash2022investigation,serbena2012internal}. We consider an idealized microstructural realization on a circular domain with radius$=0.4$, which is subject to displacement-type loading on its boundary. This microstructure realization is composed of randomly distributed crystals embedded in a glassy matrix, such that the crystals occupy $40\%$ of the volume. To generate the training and test samples, we adopted the mechanical parameters in \cite{fan2022meshfree} and employed the quasi-static linear peridynamic solid (LPS) solver to generate the high-fidelity simulation data. We train $N^{\text{train}}=\{10,40,100\}$ numbers of labelled data pairs, while validate/test the model with 40/40 samples, respectively. In this example, the specimen deformation is driven by the loading on its boundary, and hence our goal is to learn the mapping from $\fb(x):=[\ub_{BC}(x),\ab(x)]$ to $\ub(x)$, where $\ub_{BC}(x)$ stands for (padded) Dirichlet-type boundary condition, and $\ab(x)$ provides the microstructure information such that $\ab(x)=0$ if the material point $x$ is glass and $\ab(x)=0$ if $x$ is crystal. The INO-vector architecture is employed. Here, we emphasize that the domain is no longer structured and hence FNOs and MWT are not applicable. Therefore, in this example we compare the performances of GNOs and INOs.

In Table \ref{fig:vector_table}, we report our experimental results on $N^{\text{train}}=\{10,40,100\}$ samples. The proposed INOs obtain the lowest relative errors on the test dataset compared to GNOs. Furthermore, in Figure \ref{fig:vector}, we study the performance of both neural operators on translated and rotated test samples. One can see that the error from INOs is invariant, whereas GNOs errors increase with the increase of the translation and rotation ranges.

% Peridynamics

% \NL{[add 10, 40]}

% \NL{[add translational/rotational test]}
\vspace{-10pt}
\subsection{Real-world dataset: biological tissue deformation}

\vspace{-0.05in}

We now take one step further to demonstrate the performance of our method on a real-world physical response dataset not generated by solving PDEs. We consider learning the mechanical response of multiple biological tissue specimens from DIC displacement tracking measurements \citep{you2022physics}. In this example the constitutive equations and material microstructure are both unknown, and the dataset has unavoidable measurement noise. In this task, we aim to model the tissue response by learning a neural operator mapping the boundary displacement loading to the interior displacement field. We train with $N^{\text{train}}=\{10,100\}$ numbers of samples, validate with $40$ samples, and test the learned model with $4500$ samples. Since there is no known governing PDE, the PINO architecture is not applicable. Hence, we compare INOs with other three neural operator baselines (GNOs, FNOs, and MWTs). We note that experimental measurements are generally not provided on Cartesian grids. To test FNOs and MWTs, we apply a cubic spline interpolation to the
displacement field, to obtain measurements on a structured grid. The results are provided in Table \ref{fig:vector_table} and Figure \ref{fig:vector}. As one can see, INOs again perform the best. Interestingly, compared with GNOs, in this example INOs halve the test errors not only in the small data regime ($N^{\text{train}}=10$), but also in the medium data regime ($N^{\text{train}}=100$). This is possibly due to the fact that the experimental measurement noise makes the learning with a small number of samples more challenging. Therefore, this example validates the robustness of our INOs not only in a small data regime, but also on noisy real-world datasets.

% Tissue

% \NL{[add 10]}

% \YY{[number of parameters: GKN 4739330, EGKN 4746433.]}

% \NL{[Add translational/rotational test]}
\vspace{-0.05in}
\section{Conclusion}

\vspace{-0.05in}

We proposed INO to learn complex physical systems with guaranteed momentums conservation. The key is to design the network architecture that preserves the translational and rotational invariance/equivariance properties. Our approach finds a physical interpretation from a particle-based method, and only requires observed data pairs with minimal physical assumptions. 
We demonstrate with both synthetic and real-world datasets the expressivity and generalizability of the proposed INOs, and show that the guaranteed momentum conservation improves the learning efficacy, especially in small and noisy data regimes. INO is not only generalizable in handling translated and rotated datasets, but also provides improved prediction from the baseline neural operator models. Herein, we point out that our INOs represent the first conservation law guaranteed neural operator architecture, and we believe that it is a novel and promising framework applicable to many examples in physical system learning. 
%For future work, we plan to investigate the performance of INOs on time-dependent problems and their employment on long-term predictions.% constitutes an interesting future work.

\section*{Acknowledgements}

The authors would like to thank the reviewers for their careful reading and valuable suggestions that help improve the quality of the paper. 

Y. Yu, H. You, and N. Tatikola would like to acknowledge support by the National Science Foundation under award DMS-1753031 and the AFOSR grant FA9550-22-1-0197. Portions of this research were conducted on Lehigh University's Research Computing infrastructure partially supported by NSF Award 2019035.

%\newpage

\appendix

%Analogously, a translation- and rotation-equivariant INO-vector architecture can be defined. We leave further discussion in the Appendix.

\newpage
\section{Detailed Derivations}\label{app:a}

\subsection{Proof for Theorem \ref{thm:INO-scalar}}

Herein, we provide the proof for the translation- and rotation-invariant properties of INO-scalar.

\begin{proof}
Notice that $|\tilde{\fb}(Rx+g)|=|R\fb(x)|=|\fb(x)|$, hence for the lifting layer \eqref{eq:inos_p} we have
\begin{align*}
\tilde{\hb}(Rx+g,0)=\mcP[\tilde{\fb}](Rx+g)=P|\tilde{\fb}(Rx+g)|+p=P|\fb(x)|+p={\hb}(x,0).
\end{align*}
Substituting it into the iterative layer, we can prove by induction that the $j$-th layer feature function is invariant, i.e.,  $\tilde{\hb}(Rx+g,j\tau)={\hb}(x,j\tau)$, $j=0,\cdots,L$. Since the derivation for the lifting layer has already provided the base case, i.e., for $j=0$, it suffices to prove the induction step. Since $\overline{y-x}$ is frame-invariant, we have
$$\tilde{\mb}(Rx+g,Ry+g)=\kappa(\overline{y-x},\verti{\tilde{\fb}(Rx+g)},\verti{\tilde{\fb}(Ry+g)})=\kappa(\overline{y-x},\verti{{\fb}(x)},\verti{{\fb}(y)})={\mb}(x,y)$$
and hence given that $\tilde{\hb}(Rx+g,j\tau)={\hb}(x,j\tau)$, for the  $(j+1)$-th layer we have
\begin{align*}
\tilde{\hb}(Rx+g,(j+1)\tau):=&\tilde{\hb}(Rx+g,j\tau)+\tau \sigma\left(W\tilde{\hb}(Rx+g,j\tau)+\int_{{\omg}} \tilde{\mb}(Rx+g,Ry+g)\tilde{\hb}(Ry+g,j\tau) dy + c\right)\\
=&{\hb}(x,j\tau)+\tau \sigma\left(W{\hb}(y,j\tau)+R\int_{{\omg}} \mb(x,y)\hb(x,j\tau) dy + c\right)={\hb}(x,j\tau),
\end{align*}
where $\tilde{\omg}:=\{Rx+g:x\in\omg\}$, $j=0,\cdots,L-1$. Similarly, for the projection layer we obtain
$$\tilde{\mcG}[\tilde{\fb}](Rx+g)=Q_2\sigma(Q_1 \tilde{\hb}(Rx+g,L\tau)+q_1)+q_2=Q_2\sigma(Q_1 {\hb}(x,L\tau)+q_1)+q_2=\mcG[\fb](x).$$

\end{proof}

\subsection{Proof for Theorem \ref{thm:INO-vector-1}}

We now provide the detailed proof for the translation-invariant and rotation-equivariant properties for INO-vector.

\begin{proof}
Following the proof of Theorem \ref{thm:INO-scalar}, we have
$$\tilde{\hb}(Rx+g,j\tau)={\hb}(x,j\tau),\;j=0,\cdots,L,$$
$$\tilde{\mb}(Rx+g,Ry+g)=\mb(x,y).$$
Then, for the additional coordinate function, $\xb(x,j\tau)$, we have
$$\tilde{\xb}(Rx+g,0)=Rx+g=R\xb(x,0)+g.$$
Moreover, assume that $\tilde{\xb}(Rx+g,j\tau)=R{\xb}(x,j\tau)+g$, for the $(j+1)$-th layer we have
\begin{align*}
 \tilde{\xb}(Rx+g,(j+1)\tau):=&\tilde{\xb}(Rx+g,j\tau)+\tau\int_\omg (Rx+g-Ry-g)\phi(\tilde{\mb}(Rx+g,Ry+g)\tilde{\hb}(Ry+g,j\tau);w) dy\\
 =&R{\xb}(x,j\tau)+g+\tau\int_\omg R(x-y)\phi({\mb}(x,y){\hb}(y,j\tau);w) dy\\
 =&R\left({\xb}(x,j\tau)+\tau\int_\omg (x-y)\phi({\mb}(x,y){\hb}(y,j\tau);w) dy\right)+g=R{\xb}(x,(j+1)\tau)+g.
\end{align*}
Therefore, we obtain $\tilde{\xb}(Rx+g,j\tau)=R{\xb}(x,j\tau)+g$ for $j=0,\cdots,L$. Finally, we have
$$\tilde{\mcG}[\tilde{\fb}](Rx+g)=\tilde{\xb}(Rx+g,L\tau)-(Rx+g)=R\left({\xb}(x,L\tau)-x\right)=R\mcG[\fb](x).$$
\end{proof}

\subsection{A translation- and rotation-equivariant architecture}

Besides the translation-invariant and rotation-equivariant problem we have discussed in the main text, for some physical problems a translation- and rotation-equivariant INO-vector architecture would be desired. For instance, in particle-tracking problems, the current location of each molecular is of interest, which would be translation- and rotation-equivariant with the change of reference frames. In this case, one can employ the same lifting and iterative block architectures as in \eqref{eq:inov_p1}-\eqref{eq:inov_3}, and modify the projection block as
\begin{equation}\label{eq:inov_qnew}
\ub(x)= \mcQ[\xb(x,L\tau)](x):=\xb(x,L\tau).   
\end{equation}
For this architecture we have the following theorem:
\begin{theorem}[Equivariance for INO-vector] \label{thm:INO-vector-2}
The INO-vector architecture proposed in Eqs.~\eqref{eq:inov_p1}-\eqref{eq:inov_3} and \eqref{eq:inov_qnew} is translation- and rotation-equivariant. That means, when translating the reference frame by $g\in\real^d$ and rotating it by an orthogonal matrix $R\in\real^{d\times d}$, the following property holds true:
$$\tilde{\mcG}[\tilde{\fb};\theta](Rx+g)=R\tilde{\mcG}[\fb;\theta](x)+g \text{ ,}$$
where $\tilde{\fb}(Rx+g):=R\fb(x)$.
\end{theorem}
\begin{proof}
The proof can be trivially obtained following the same argument as in Theorem \ref{thm:INO-vector-1}.
\end{proof}

\section{Problem definitions and network settings}\label{app:b}

We note that, in order to demonstrate the full expressivity of INO and carry out fair comparisons with GNO, MWT, and FNO, all the involved integral operators are evaluated over the entire domain. However, in practical applications, one can strike a balance between computational time and accuracy by evaluating the integral over a ball of a smaller radius. Moreover, to prevent overfitting, an early stopping scheme is employed: the training process is stopped if the validation loss does not drop in 60 epochs, and validation is performed only when the train loss is improved over the last saved model. For fair comparison, the hyperparameters are tuned for each method, including the learning rates (tuned in the range $[1e-2, 1e-4]$), decay rates (tuned out of $\{0.5, 0.7, 0.9\}$), and regularization coefficients (tuned in the range $[1e-2, 1e-5]$). In PINOs, we further tune the penalty coefficient for the PDE loss in the range $[0.1,10]$. All the models are trained up to a total of 2000 epochs, and the learning rate is decayed at the interval of every 50 epochs. In all the examples, the dimension of representation is set to $d_h=64$, and the kernel $\kappa$ is modeled by a 3-layer MLP with width $(n, 512, 1024, d_h^2)$, where $n$ is the number of kernel arguments for each architecture. Finally, the output $\ub(x)\in\real^{d_u}$ at each point $x$ is obtained via a projection layer in the form of a 2-layer MLP with width $(d_h, 2d_h, d_u)$.

\begin{table}[h!]
    \centering
    {\small\begin{tabular}{|c|c|c|c|c|}
    \hline 
        {Model, Dataset, Resolution} & \multicolumn{4}{c|}{Number of training samples} \\
        \cline{2-5}
        & 10 & 20 & 40 & 100 \\
        \hline \hline
         FNO, train, $16\times 16$& %0.045\%  & 
         0.153\% & 0.715\% & 0.906\% & 0.216\%\\
         FNO, test, $16\times 16$& %46.15\% & 
         14.29\% & 8.582\% & 5.235\% & 2.510\% \\
         FNO, test, $31\times 31$& %43.37\% & 
         14.75\% & 10.69\% & 9.072\% &  8.244\% \\
\hline         
         GNO, train, $16\times 16$&  %3.652\%  & 
         2.117\% & 4.730\% & 2.002\% & 0.872\%\\
         GNO, test, $16\times 16$&  %10.978\%  & 
         7.708\% & 6.536\% & 4.389\% & 1.966\%\\
         GNO, test, $31\times 31$&  %11.046\%  &
         8.167\% & 7.659\% & 7.579\% & 5.842\%\\
\hline         
         INO, train, $16\times 16$&  %5.214\%  & 
         1.414\% & 1.580\% & 1.068\% & 0.500\%\\
         INO, test, $16\times 16$&  %11.010\%  & 
         6.145\% & {\bf 4.922\%} & {\bf 3.110\%} & {\bf 1.182\%}\\
         INO, test, $31\times 31$&  %11.140\%  &
         {\bf 6.900\%} & {\bf 6.345\%} & {\bf 4.693\%} & {\bf4.167\%}\\
\hline         
         norm-INO, train, $16\times 16$&  1.008\% & - & - & 0.627\%\\
         norm-INO, test, $16\times 16$& {\bf 4.598\%} & - & - & 1.273\%\\
\hline         
         PINO, train, $16\times 16$&  3.105\% & - & - & 0.468\%\\
         PINO, test, $16\times 16$&  9.652\% & - & - & 2.560\%\\
\hline 
         aug-GNO $N^{aug}$=20, train, $16\times 16$&  10.036\% & - & - & -\\
         aug-GNO $N^{aug}$=20, test, $16\times 16$&   18.050\% & - & - & -\\
\hline 
         aug-GNO $N^{aug}$=40, train, $16\times 16$&   1.822\% & - & - & -\\
         aug-GNO $N^{aug}$=40, test, $16\times 16$&  8.850\% & - & - & -\\
\hline 
         MWT, train, $16\times 16$&   0.688\% & 0.962\% & 0.168\% & 0.129\% \\
         MWT, test, $16\times 16$&  10.56\% & 7.406\% & 5.473\% & 3.420\% \\
\hline 
         EGNN, train, $16\times 16$&  16.56\% & - & - & 7.710\%\\
         EGNN, test, $16\times 16$&  19.67\% & - & - & 9.550\%\\
         \hline 
    \end{tabular}}
    \caption{Results for the sub-surface flow problem, where bold numbers highlight the best method for the test datasets on original-resolution ($16\times 16$ grids) and refined-resolution ($31\times 31$ grids), respectively.}
    \label{tab:darcy_results}
\end{table}

\begin{table}[]
    \centering
    \begin{tabular}{|c|c|c|c|c|c|c|}
    \hline 
         model & INO & GNO & FNO & PINO & MWT & EGNN \\
         \hline
         nparams & 4,739,201 & 4,740,353 & 4,219,841 & 4,219,841 & 4,565,185& 4,677,156   \\
         \hline 
    \end{tabular}
    \caption{Total number of parameters of each model for the sub-surface flow problem.}
    \label{tab:darcy_para}
\end{table}

\begin{figure}[!t]\centering
 \includegraphics[width=.99\linewidth]{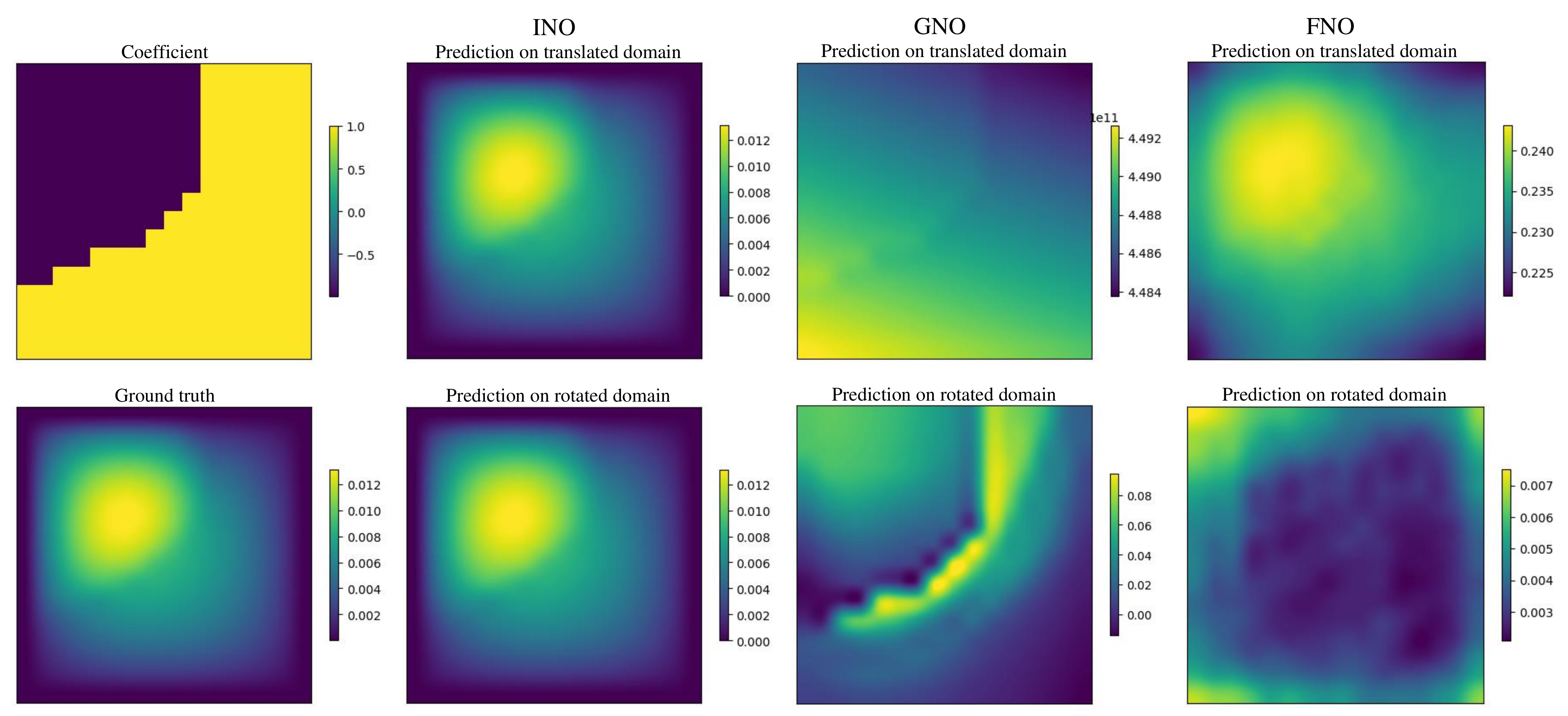}
 \caption{Demonstration of the predictions of INO, GNO, and FNO on randomly translated and rotated domains.}
 \label{fig:plot_trans_rot_pred}
\end{figure}

\begin{figure}[!t]\centering
\includegraphics[width=.9\linewidth]{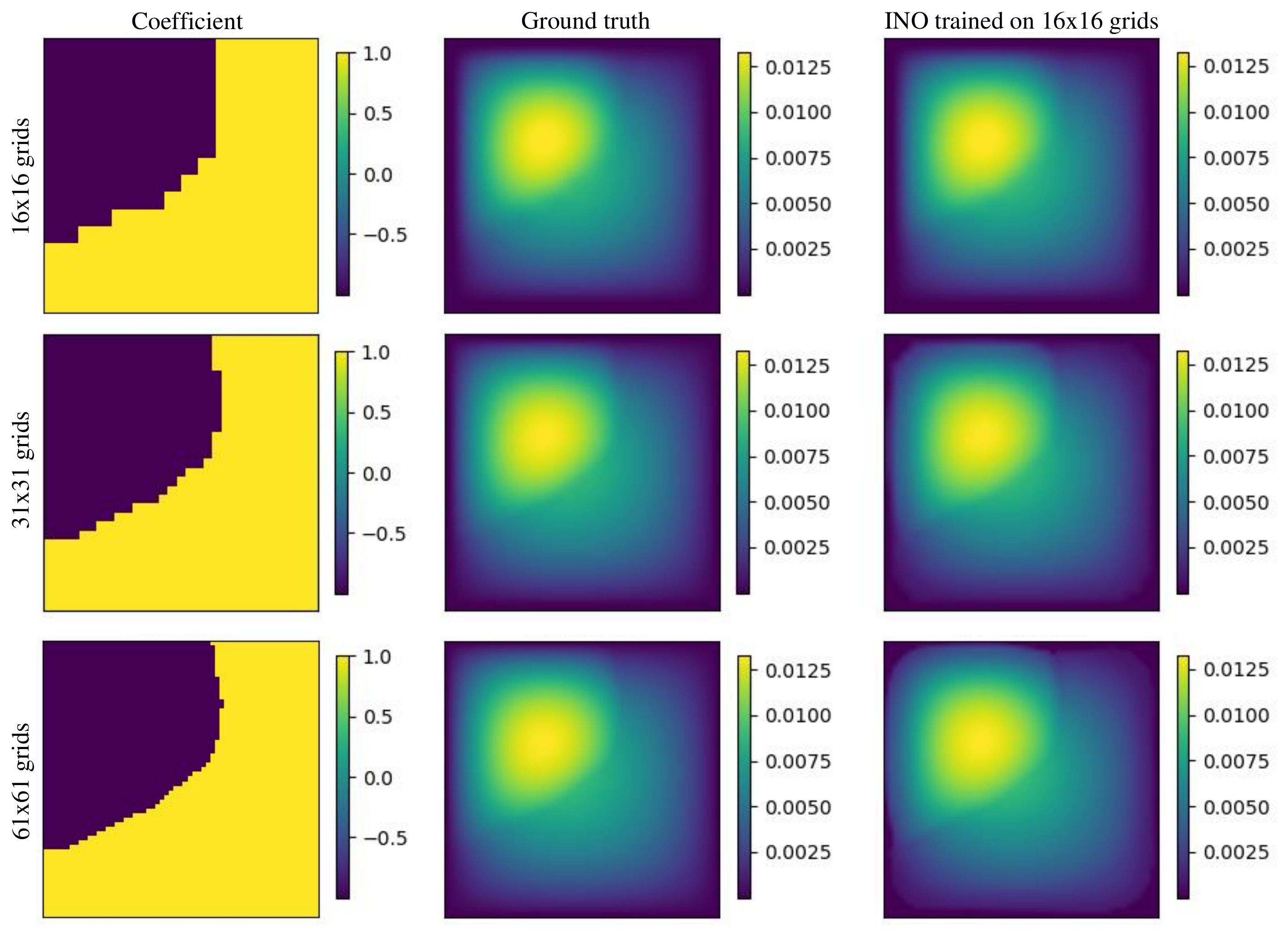}
 \caption{Demonstration of the cross-resolution predictability of INO trained on $16\times16$ grids and predicted directly on $31\times31$ and $61\times61$ grids.}
 \label{fig:plot_cross_resolution}
\end{figure}

\subsection{Synthetic dataset: sub-surface flow}

The detailed numerical results for ablation study and empirical experiments (displayed in Figure \ref{fig:darcy}) are provided in Table~\ref{tab:darcy_results}. The number of trainable parameters are listed in Table~\ref{tab:darcy_para}. Furthermore, a demonstration of the predictions of INO, GNO, and FNO on randomly translated and rotated datasets is displayed in Figure \ref{fig:plot_trans_rot_pred}. In order to demonstrate the resolution independence nature of INO, an example is illustrated in \ref{fig:plot_cross_resolution} where INO is trained on $16\times16$ grids and predicted directly on $31\times31$ and $61\times61$ grids.

\subsection{Synthetic dataset: glass-ceramics deformation}

\begin{figure}[!t]\centering
 \includegraphics[width=.99\linewidth]{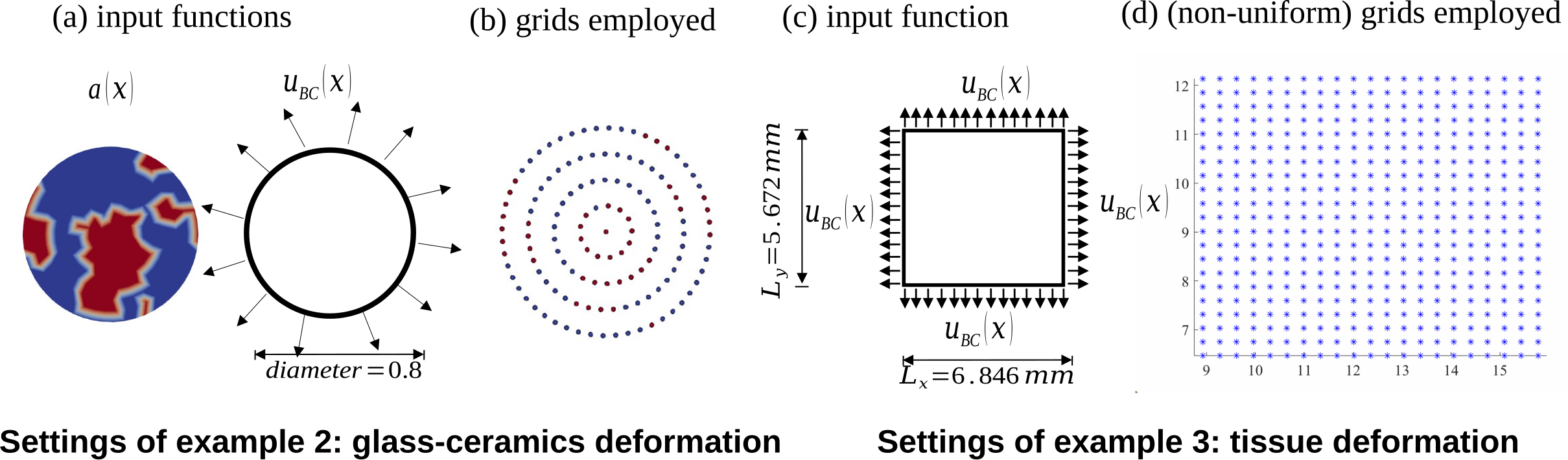}
 \caption{Input functions and (non-uniform) grids employed in examples 2 and 3.}
 \label{fig:darcy_loss}
\end{figure}

Following \cite{fan2022meshfree}, we model the glass-ceramic material deformation with the linear peridynamic solid (LPS) model:
\begin{equation}\label{eq:nonlocElasticity}
\begin{aligned}
    \mcL_\delta \ub:=&-\frac{1}{m(\delta)}  \int_{B_\delta (x)} \left(\lambda(x,y)- \mu(x,y)\right) K(\left|y-x\right|) \left(y-x \right)\left(d(x) + d(y) \right) dy\\
  &
  -  \frac{8}{m(\delta)}\int_{B_\delta (x)} \mu(x,y) K(\left|y-x\right|)\frac{\left(y-x\right)\otimes\left(y-x\right)}{\left|y-x\right|^2}  \left(\ub(y) - \ub(x) \right) dy = 0 \text{ ,}\quad \text{ for }x\in\omg \text{ ,}\\
  \ub(x):=&\ub_{BC}(x) \text{ ,} \quad \text{ for }x\in\omgbb \text{ ,}
   \end{aligned}
\end{equation}
where the nonlocal dilatation $d(x)$ is defined as
\begin{equation}\label{eqn:oritheta}
d(x):= \frac{1}{m(\delta)}\int_{B_\delta (x)} K(\left|y-x\right|) (y-x)\cdot \left(\ub(y) - \ub(x) \right)dy,\text{ for }x\in\omg\cup\omgb \text{ .}
\end{equation}
Here, $\omgb$ and $\omgbb$ are the one-layer and two-layer nonlocal boundaries, respectively, which will be defined later. The kernel function $K$ and nonlocal volume $m(\delta)$ are defined as 
\begin{equation} \label{eqn:kernel}
    K(|y-x|) = \frac{1}{|y-x|}\text{ ,}\quad m(\delta):=\int_{B_{\delta}(\bm{0})} K(|y|) |y|^2 d y = \frac{2\pi \delta^3}{3}\text{ .}
\end{equation}
$\mu(x,y)$, $\lambda(x,y)$ are the averaged shear modulus and Lam{\'{e}}  first parameters of material points $x$ and $y$.

% \begin{table}[]
% \center
% \begin{tabular}{|c|cccc|}
% \hline
% &Young's modulus& Poisson ratio& Fracture energy& Fracture Toughness\\
% \hline
% Glass&$E_1=$80 $GPa$ &0.25&$G_1=$6.59 $J/m^2$ & 0.75 $MPa\cdot m^{0.5}$\\
% Crystal&$E_2=$133 $GPa$&0.25&$G_2=$86.35 $J/m^2$ & 3.5 $MPa\cdot m^{0.5}$\\
% \hline
% \end{tabular}
% \caption{Material properties used in the glass-ceramics experiment.}
% \label{tab:glass}
% \end{table}

To model the glass-ceramic specimen, we consider a circular domain with the same material parameters as provided in \cite{fan2022meshfree}. The microstructure realization is randomly generated such that the volume ratio between crystal and glass is around 2:3. To generate the displacement under different boundary conditions, we consider the horizon size $\delta = 0.3$, the nonlocal domain as $\omg = \{x:|x| \leq 0.4 \}$, and the nonlocal boundary regions as $\omgb = \{ x:0.4<|x| \leq 0.7\}, \omgbb = \{x: 0.4<|x| \leq 1.0\}$. For each sample, the nonlocal boundary condition $\ub_{BC}(x)$ is generated from a Gaussian random field with $\mathbf{G}(x) = Re \Bigg(\begin{bmatrix}
\sum_{k_1,k_2} \xi_{k} \exp(i 2\pi k \cdot x /D)\mcU[0,1] \\
\sum_{k_1,k_2} \xi_{k} \exp(i 2\pi k \cdot x /D)\mcU[0,1] 
\end{bmatrix} \Bigg)$, where $k=(k_1,k_2)$, $D = 2.8$, $\xi_{k} = (k_1^2+k_2^2)^{-5/4}$, $\mcU[0,1]$ represents the uniform distribution on $(0,1)$, and $k_1,k_2 \in \{-15,-14,\dots, 13\}$. The Gaussian random field $\mathbf{G}(x)$ is created on a square domain $[-1.4,1.4] \times [-1.4,1.4]$ with structured grids, and the nonlocal boundary condition $\ub_{BC}$ is obtained by cubic interpolation on unstructured grids within the region $\omgbb$. The corresponding high-fidelity displacement field $\ub(x)$ is generated employing the meshfree solver as described in \cite{fan2022meshfree}. %, and the Dirichlet-type boundary condition is defined as $\ub_{BC}:=\ub(x), x\in \partial \omg$.
Here, we point out that in this example the domain shape and the grids are not structured, hence the FNO and MWT models are not applicable and we focus on the comparison between GNOs and INOs.

\subsection{Real-world dataset: biological tissue 
deformation}

% \begin{figure*}[!ht]
% \centering
% \includegraphics[width=1.0\columnwidth]{./figures/setting2.png}
% \caption{Biological tissue test setup for generating the real-world dataset: (a) an image of the speckle-patterned porcine tricuspid valve anterior leaflet (TVAL) specimen subject to biaxial stretching, (b) schematic of the specimen subject to Dirichlet-type boundary conditions as the corresponding numerical setting of (a), and (c) the DIC (digital image correlation) tracking grid where the points are not strictly uniformly spaced.}
% \label{fig:dicsetup}
% \end{figure*}

We briefly introduce the experimented specimen and the corresponding biological tissue dataset. In this example, we employ the opensource dataset provided in \url{https://github.com/fishmoon1234/IFNO-tissue}, where a tricuspid valve anterior leaflet tissue with an effective testing area of $9\times9$\,mm was mounted on a biaxial testing device. Then, various loadings were applied on the boundary of this tissue specimen, and the corresponding displacement was recorded for each material point 
%Specifically, we followed our previous biaxial testing process, including acquiring a healthy porcine heart and retrieving the TVAL \citep{you2022physics}. 
%The leaflet tissue was then cut into a square-shaped specimen. Subsequently, a speckling pattern was imposed to the tissue surface, which was then mounted onto a biaxial testing device with an effective testing area of $9\times9$\,mm for the following tissue property characterizations (Figure  \ref{fig:dicsetup}(a)). Throughout the test, a CCD camera was used to capture images of the tested specimen, and the load cell readings and actuator displacements were recorded at a frequency of 5\,Hz. 
by employing the digital image correlation (DIC) toolkit of the BioTester's software. The unstructured coordinate locations of the DIC-tracked grids were directly used to train the proposed INO and reference GNO models. To create a structured grid for FNOs and MWTs, we further applied a cubic spline interpolation to the
displacement field on a structured $21\times 21$ node grid. The dataset consists of a total of $26,523$ time instants (samples), which is further divided into $22,023$ for training and $4,500$ for testing. In our example, we focus on the small training data regime, by randomly select $10$ or $100$ numbers of samples for the purpose of training, then validate the model on all $4,500$ test samples and provide the averaged relative error for displacement fields.

\end{document}